\documentclass[11pt]{article}

\usepackage{acl}

\usepackage{times}
\usepackage{latexsym}
\usepackage[T1]{fontenc}
\usepackage[utf8]{inputenc}
\usepackage{microtype}
\usepackage{inconsolata}
\usepackage{graphicx}
\usepackage{dblfloatfix}
\usepackage{algorithm}
\usepackage{algpseudocode}
\usepackage{amsmath}
\usepackage{amssymb}
\usepackage{booktabs}
\usepackage{colortbl}
\usepackage{xcolor}
\usepackage{textcomp}

\title{\textsc{VICT}: Verifier-Instrumented Credit Tracing for Long-Horizon LLM Agent Reinforcement Learning}

\author{
\textbf{Pengcheng Li}\textsuperscript{1} \quad
\textbf{Zhengyang Zhang}\textsuperscript{1} \quad
\textbf{Dongxu Zhang}\textsuperscript{2} \\
\textbf{Sui Huang}\textsuperscript{3} \quad
\textbf{Shaohua Ma}\textsuperscript{1,*} \\
{\normalfont \textsuperscript{1}Tsinghua University} \\
{\normalfont \textsuperscript{2}Xi'an Jiaotong University} \\
{\normalfont \textsuperscript{3}Jiaxing Nanhu University} \\
{\normalfont\small \textsuperscript{*}Corresponding author. Email: \texttt{lipc24@mails.tsinghua.edu.cn}}
}

\begin{document}
\maketitle

\begin{abstract}
Fine-grained credit assignment is a central challenge in reinforcement learning for long-horizon LLM agents. Standard objectives often train from programmatically verifiable terminal rewards by broadcasting each sparse outcome to every action in a trajectory. Existing methods typically seek finer credit from the rollout side, constructing auxiliary trajectory signals or additional comparisons to estimate action importance. Although useful, these approaches still treat the verifier that judged success as a scalar reward, discarding its internal task structure. Our key insight is that many verifiable tasks already encode the relevant checks inside their terminal verifier. We propose \textsc{VICT} (Verifier-Instrumented Credit Tracing), a training-time interface that exposes executable or evidence-backed atoms and traces them back to actions through dependency-valid proof edges. \textsc{VICT} redistributes group-relative advantage only along those edges, shifting credit assignment from rollout-side inference to verifier-side tracing. It preserves the original terminal reward, abstains when evidence is incomplete or ambiguous, and changes only the training-time advantage tensor, requiring no learned critic, process labels, branch rollouts, or inference-time verifier access. On ALFWorld and WebShop, \textsc{VICT} improves substantially over outcome-only training and achieves strong performance alongside recent fine-grained credit methods; ablations rule out dense atom rewards, final-commit credit, temporal proximity, and sparsity as sufficient explanations.

\end{abstract}

\section{Introduction}
\label{sec:introduction}

Large language models (LLMs) are increasingly trained as agents that interact with external environments over multiple turns, where success depends on making a sequence of decisions rather than producing a single response~\citep{yao2023react,schick2023toolformer,shridhar2020alfworld,yao2022webshop}. Reinforcement learning (RL) is a natural fit for such agents because many tasks expose verifiable outcome rewards, and group-based methods such as RLOO and GRPO make this setting practical by estimating advantages from rollout groups without a learned critic~\citep{ahmadian2024back,shao2024deepseekmath}. Yet in long-horizon interaction, a sparse terminal reward is usually assigned to every action in the trajectory, even though only a few decisions may determine success or failure.

The central challenge is therefore not only that rewards are sparse, but that terminal outcomes erase the reason why a trajectory succeeded or failed~\citep{zhang2026spark,zhang2026pointcot}. To recover finer credit, recent methods construct comparison units from repeated states, trajectory graphs, semantic proximity, hindsight or process-level feedback, and dynamic branches~\citep{feng2025group,li2026salt,fang2026proxmo,tan2026hcapo,lightman2023let,xi2025agentprm,ji2026treegrpo,dong2025arpo,wu2026spark}. These methods improve over uniform trajectory-level credit, but they typically recover credit from rollout-side proxy signals rather than from the verifier that defined the terminal outcome. State or trajectory similarity can miss verifier-relevant history or merge states with different task-critical facts; hindsight or process feedback can depend on auxiliary judgments; and branching buys information with extra sampling cost. In all cases, the actual rule that judged the task remains mostly outside the credit assignment mechanism.

\begin{figure}[t]
    \centering
    \includegraphics[width=\columnwidth]{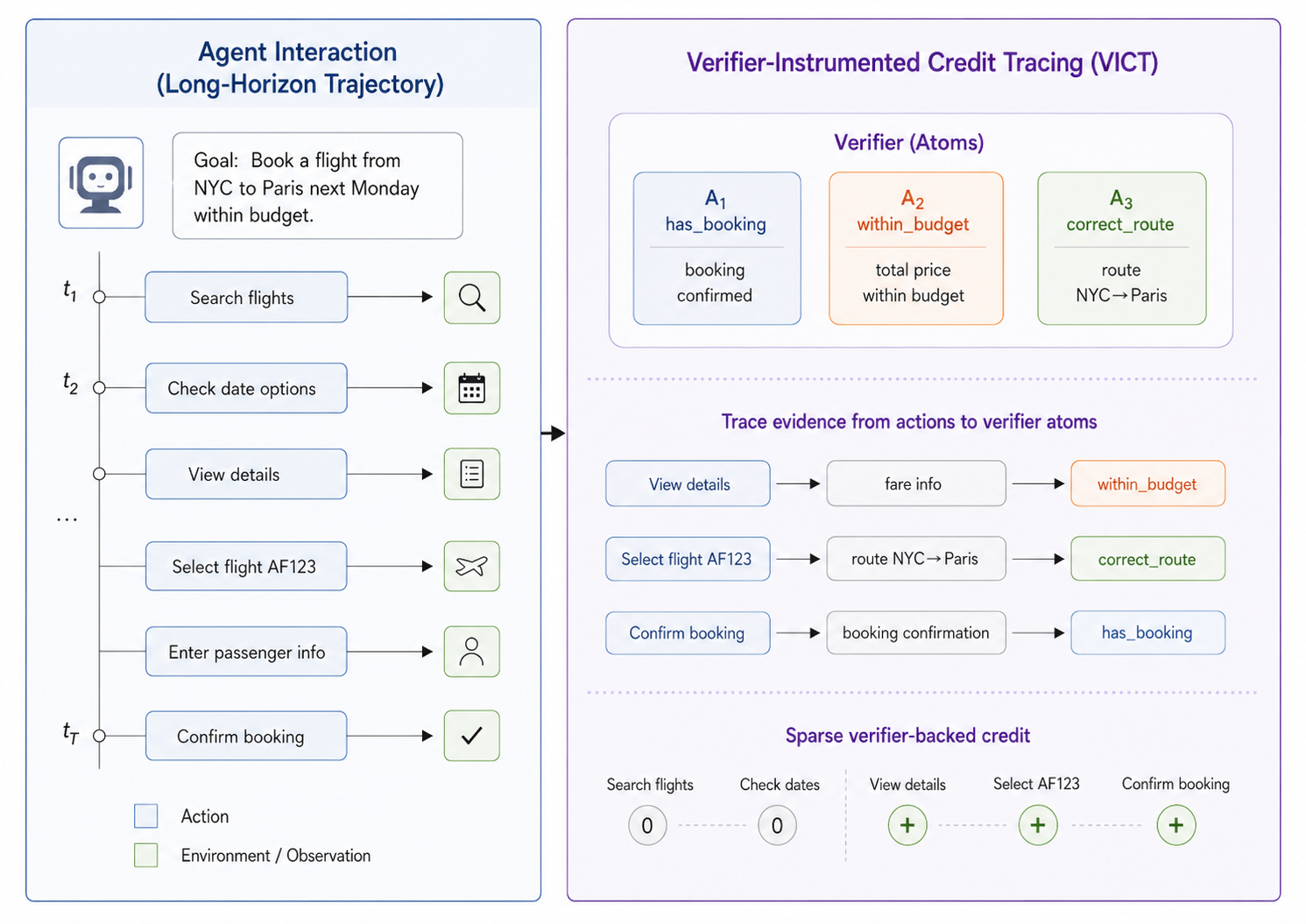}
    \caption{Overview of verifier-instrumented credit tracing. A long-horizon flight-booking trajectory is checked by verifier atoms such as booking completion, budget satisfaction, and route correctness; \textsc{VICT} traces evidence from concrete actions to these atoms and assigns sparse verifier-backed credit only to supported steps.}
    \label{fig:intro}
    \vspace{-0.6em}
\end{figure}

This limitation suggests a different source of credit: instead of reconstructing action importance from the scalar outcome, we can inspect how the outcome was produced. In many verifiable agent tasks, the terminal reward is computed by a programmatic verifier that checks concrete facts, such as required state changes, forbidden operations, observed evidence, and final commitments. Prior work on reward shaping, reward machines, and reward decomposition shows that reward structure can provide useful learning signal~\citep{ng1999policy,toroicarte2018reward,juozapaitis2019explainable}; here, the structure already exists inside the terminal verifier. Standard RL interfaces discard these checks and expose only the final number. This motivates our central question: can the terminal verifier be used not merely as an outcome oracle, but as a training-time credit tracer?

We propose \textsc{VICT}, Verifier-Instrumented Credit Tracing, which turns instrumentable terminal verifiers into sparse, auditable action-level credit. \textsc{VICT} instruments a programmatic verifier into executable or evidence-backed atoms, links these atoms to observable trajectory evidence, and redistributes group-relative advantage only through verified action-to-atom proof edges. In this paper, verifier-backed credit means an eligibility guarantee rather than a causal proof: a correction is emitted only when the verifier interface conforms, a dependency-valid core identifies the relevant atom, and a fixed witness relation links that atom to an action. The method keeps the original terminal reward as the outcome anchor, changes only the training-time advantage tensor, and abstains when no reliable verifier proof exists. It does not assume automatic decompilation of arbitrary black-box judges; the verifier interface is an explicit engineering object whose conformance and cost must be audited.

This framing shifts the problem from inventing intermediate rewards to exposing the structure of existing verifiers. We evaluate \textsc{VICT} on long-horizon verifiable agent benchmarks against outcome-only and fine-grained RL baselines. On ALFWorld and WebShop, \textsc{VICT} improves substantially over GRPO and remains competitive with recent fine-grained methods, while $\tau$-bench provides suggestive service-agent validation under a different backbone and protocol. We also audit the verifier interface through reconstruction, mutation conformance, eligibility, coverage, sparsity, abstention, core-size, and cost diagnostics~\citep{sun2026tri}. Our contributions are:

\begin{itemize}
    \item We formulate verifier instrumentation as a training-time credit interface for programmatically verifiable LLM agent RL.
    \item We introduce dependency-core attribution and proof-edge-constrained advantage correction with an explicit eligibility invariant.
    \item We provide experiments, ablations, and interface diagnostics showing when verifier-backed credit improves over outcome-only training and recent credit-assignment baselines.
\end{itemize}

\section{Related Work}
\label{sec:related_work}

\subsection{Reinforcement Learning for LLM Agents}
\label{sec:rw_llm_agent_rl}

LLM agents extend language models from static generation to interactive decision making, where policies must use observations, tools, and external feedback over multiple turns. Prompting, tool-use training, and supervised agent tuning have enabled strong zero-shot and imitation-based agents~\citep{yao2023react,schick2023toolformer,zeng2024agenttuning}. Reinforcement learning further allows agents to optimize verifiable outcomes directly, and has been applied to web, embodied, search, tool-use, and application-control settings~\citep{yao2022webshop,shridhar2020alfworld,jin2025search,trivedi2024appworld,yao2024tau}.

Most scalable LLM RL pipelines build on policy-gradient objectives~\citep{li2025emo}. PPO-style training uses a value model, while RLOO and GRPO avoid a learned critic by normalizing rewards within rollout groups~\citep{schulman2017proximal,ahmadian2024back,shao2024deepseekmath}. This critic-free design is attractive for long-context agent trajectories, but it still leaves the key question of how a terminal outcome should be assigned to individual actions. \textsc{VICT} keeps the group-based optimization setting and changes only the training-time advantage signal.

\subsection{Credit Assignment under Sparse Outcome Rewards}
\label{sec:rw_sparse_credit}

Sparse terminal rewards make long-horizon agent training inefficient because the same trajectory-level advantage is often broadcast to every action. Recent and concurrent methods refine this signal by changing the comparison unit. GiGPO compares actions from repeated environment states, SALT uses trajectory graphs to distinguish shared and divergent steps, ProxMO replaces hard state groups with semantic proximity, and HCAPO estimates action utility through hindsight reasoning~\citep{feng2025group,li2026salt,fang2026proxmo,tan2026hcapo}. Several of these references are recent preprints. Hierarchical and subgoal-based methods assign credit at coarser temporal abstractions~\citep{peng2026hiper,wang2026subgoal,xue2026strata}.

Another line changes the rollout distribution rather than the credit rule. Tree- and branch-based methods sample continuations from shared prefixes or uncertain decision points to expose local preferences~\citep{ji2026treegrpo,dong2025arpo,zhao2026branpo,wu2026spark}. These approaches are effective when useful comparisons can be created from states, prefixes, branches, or prompted hindsight. \textsc{VICT} targets a complementary source of supervision: the structure of the terminal verifier itself. It assigns extra credit only when an action can be linked to a verifier atom through an observable proof edge.

\subsection{Structured Verifier Supervision and Rubric-Based Evaluation}
\label{sec:rw_structured_verifiers}

Prior RL work has shown that structure inside the reward can be useful for learning and interpretation. Return redistribution, potential-based shaping, reward machines, and reward decomposition expose delayed or composite rewards in more informative forms~\citep{arjona2019rudder,ng1999policy,toroicarte2018reward,juozapaitis2019explainable}. Process supervision and PRMs similarly provide step-level feedback, but they typically require labels, learned reward models, or search-generated supervision~\citep{uesato2022process,lightman2023let,luo2024omegaprm,xi2025agentprm}. \textsc{VICT} is best viewed as verifier-grounded advantage redistribution: it changes the policy-gradient update signal, but it uses the programmatic verifier that already defines the environment outcome rather than introducing a separate progress reward or reward model.

Rubric and criteria-based evaluation decomposes holistic judgments into interpretable dimensions, and recent work uses such criteria as optimization signals~\citep{xie2026autorubric,yu2025dece,gunjal2025rubrics}. However, natural-language atomic decomposition can be unreliable when criteria are not executable or evidence-grounded~\citep{zhang2026atomic}. \textsc{VICT} therefore treats verifier decomposition as a constrained instrumentation problem: atoms must be executable or evidence-backed, reconstruct the terminal verifier, and attach to actions through proof edges. This makes its advantage correction auditable and verifier-aligned without turning rubric criteria into dense rewards.

\section{Method}
\label{sec:method}

\subsection{Problem Setup and Design Goal}
\label{sec:method_setup}

We consider long-horizon interactive RL where a language-agent policy $\pi_\theta$ solves a task instance $x$ through observations and actions. A rollout is
\begin{equation}
\tau_i=(h_{i,0},a_{i,0},h_{i,1},a_{i,1},\ldots,h_{i,T_i})
\label{eq:rollout}
\end{equation}
where $h_{i,t}$ is the history before action $a_{i,t}$, and the environment returns a terminal verifier score $R_i=V_x(\tau_i)$. Standard group-based optimization yields a task-level base advantage $A_i^{\mathrm{base}}$ and usually broadcasts it to all actions, so it cannot distinguish state-changing, evidence-revealing, violating, or merely co-occurring actions.

\textsc{VICT} keeps the outcome signal but changes the verifier-optimizer interface. It treats $V_x$ as an instrumentable audit interface: verifier atoms expose checked facts, dependencies expose valid combinations, and trajectory witnesses decide which actions may receive additional credit. Rollout collection and inference-time policy behavior are unchanged.

The central object is a verifier-derived credit trace:
\begin{equation}
\Gamma_i=
(\mathbf z_i,G_i,C_i^\star,\mathcal P_i),\quad
\mathbf z_i=(z_{i,j,t})_{j,t}
\label{eq:vict_trace}
\end{equation}
Here $\mathbf z_i$ is the temporal atom trace, $G_i$ is an action-to-atom proof graph, $C_i^\star$ is a dependency-closed verifier core, and $\mathcal P_i$ stores proof records. The trace supports verifier faithfulness, proof-edge eligibility, and clipped advantage correction anchored by the original outcome.

\begin{figure*}[t]
    \centering
    \includegraphics[width=\textwidth]{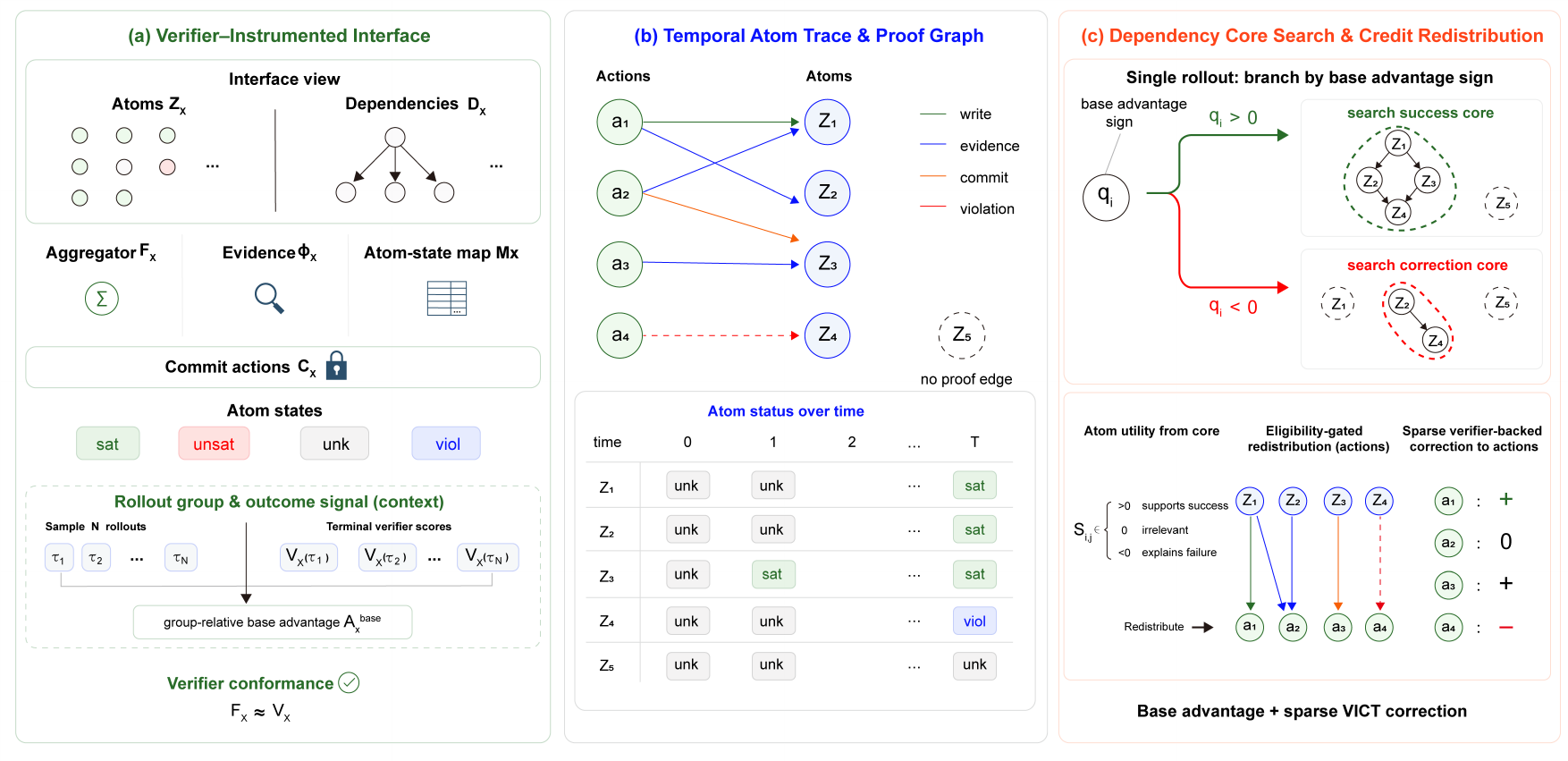}
\caption{\textsc{VICT} pipeline. The verifier is first exposed as atoms, dependencies, evidence extractors, and commit predicates; trajectories are then converted into proof graphs that link actions to verifier atoms. Dependency cores estimate which atoms matter to success or failure, and the final step redistributes normalized credit only to proof-supported actions.}
    \label{fig:method}
    \vspace{-0.6em}
\end{figure*}

\subsection{Verifier-Instrumented Interface}
\label{sec:verifier_interface}

For each task instance, \textsc{VICT} represents the verifier with executable atoms, an atom-to-score aggregator, dependency rules, state or evidence bindings, evidence extractors, and commit predicates. Each atom has a typed status $z_{i,j,t}\in\{\mathrm{sat},\mathrm{unsat},\mathrm{unk},\mathrm{viol}\}$. The aggregator $F_x$ maps terminal atom assignments to the verifier score, $D_x$ encodes dependency validity, $M_x$ links atoms to state or evidence variables, $\Phi_x$ extracts observable evidence for write and reveal witnesses, and $C_x$ identifies commit actions. We require the dependency closure $\mathrm{cl}_{D_x}$ to be deterministic; otherwise the affected atoms are not used for credit.

The interface is accepted only if it passes verifier conformance. Let $\bar{\mathbf z}(\tau)$ be the terminal atom assignment and $\mathcal S_x=\mathcal T_x\cup\mathfrak M_x(\mathcal T_x)$ include validation rollouts and verifier-relevant mutations. Define
\begin{equation}
\epsilon_x^{\mathrm{conf}}
=
\max_{\tilde{\tau}\in\mathcal S_x}
\left|F_x(\bar{\mathbf z}(\tilde{\tau}))-V_x(\tilde{\tau})\right|
\label{eq:verifier_conformance}
\end{equation}
The interface is accepted when $\epsilon_x^{\mathrm{conf}}\leq\eta_x$. For exact verifiers $\eta_x=0$; for graded verifiers the tolerance is fixed before training. When conformance fails for a task or atom, \textsc{VICT} abstains from using that atom. The adapter may reveal atoms from final-state differences, verifier branches, rule predicates, or evidence extractors, but it cannot introduce policy-visible subgoals or new preferences.

Concrete domain instantiations and a WebShop proof trace are provided in Appendices~\ref{app:verifier_interface_construction} and~\ref{app:illustrative_traces}.

\subsection{Credit Trace and Core Attribution}
\label{sec:credit_trace_core}

Given a trajectory and a conforming interface, \textsc{VICT} evaluates $\{z_{i,j,t}\}_{j,t}$ and constructs $G_i=(A_i,Z_i,E_i)$. Edges are created only when a witness predicate certifies that an action wrote, revealed, committed, or violated an atom. Let $\Delta_{i,t}(v)$ denote a change in extracted evidence, $\Phi_x(h_{i,t+1})[v]\neq\Phi_x(h_{i,t})[v]$. Two primitive witnesses are
\begin{equation}
\begin{aligned}
W_{\mathrm{write}}
&=\mathbf{1}\{\exists v\in M_x(z_j):\Delta_{i,t}(v)=1\}\\
W_{\mathrm{commit}}
&=\mathbf{1}\{a_{i,t}\in C_x,\ j\in\mathrm{scope}_x(a_{i,t})\}
\end{aligned}
\label{eq:witness_primitives}
\end{equation}
The commit scope includes the verifier atoms finalized by the action. Other witnesses use the same pattern: they must be computed from logged history, extracted evidence, and verifier metadata, rather than from a learned relevance model. The proof graph contains each action-atom-relation triple whose witness predicate is true, and $\mathcal P_{i,t,j}$ stores the matched relations together with core and conformance tags.
Terminal-only atoms with no reliable writer or commit action create no edge; unsupported attribution is treated as missing information.

The proof graph determines where credit may go, but not how much an atom matters to the terminal verifier. Let $\bar{\mathbf z}_i=(z_{i,1,T_i},\ldots,z_{i,m,T_i})$ and let $\epsilon_q$ be the pre-specified tie margin in base-advantage units. Rather than introducing a separate success/failure heuristic, \textsc{VICT} uses the sign of the same group-normalized base advantage used by the underlying optimizer:
\begin{equation}
q_i =
\begin{cases}
+1, & A_i^{\mathrm{base}}>\epsilon_q,\\
-1, & A_i^{\mathrm{base}}<-\epsilon_q,\\
0, & |A_i^{\mathrm{base}}|\leq\epsilon_q .
\end{cases}
\label{eq:base_adv_direction}
\end{equation}
When $q_i=0$, the verifier correction abstains. Positive-sign rollouts search for success cores whose removal would lower the verifier score, while negative-sign rollouts search for correction cores whose repair would raise it. This uses the optimizer's own preference signal rather than a separate median heuristic. The functional form of $\rho_i$ is fixed before training, while its value is computed from the current rollout group's base preference magnitude (Appendix~\ref{app:attribution_details}). Counterfactuals are verifier-assignment edits, not new environment rollouts, and ambiguous dependency projections trigger abstention. For readability, write $\Delta_x(C)=\Delta_x(C;\bar{\mathbf z}_i,q_i)$. The verifier displacement is
\begin{equation}
\Delta_x(C)
=
q_i\!\left[F_x(\bar{\mathbf z}_i)-F_x\!\left(\mathrm{cf}^{q_i}_{D_x}(\bar{\mathbf z}_i,C)\right)\right]
\label{eq:dependency_core}
\end{equation}
The core $C_i^\star$ is a compact dependency-closed set returned by budgeted greedy search such that $\Delta_x(C_i^\star)\geq\rho_i$ under budget $B$; details and local-minimality scope are in Appendix~\ref{app:budgeted_greedy_search}. If no valid core reaches $\rho_i$, the budget is exhausted, or the displacement lies inside the conformance uncertainty band, \textsc{VICT} abstains. Core atoms receive leave-one-out marginals, while atoms outside the core receive zero marginal.

\subsection{Proof-Edge-Constrained Advantage Optimization}
\label{sec:advantage_optimization}

Core marginals are normalized within the same-task rollout group $\mathcal G_x$ used by the base group optimizer; historical rollouts are not mixed into the statistics. We use a robust group scale defined in Appendix~\ref{app:robust_normalization}. If the scale is zero or the group has too few supported rollouts, the normalized value is set to zero. The normalized signal is used only if the atom has both core membership and proof support. If conformance, core search, or proof support fails, the corresponding eligibility mask is zero, so \textsc{VICT} falls back to $A_i^{\mathrm{base}}$. Let $Z_{i,j}$ denote the total witness weight for atom $j$ plus smoothing. The eligible core signal is redistributed through the proof graph:
\begin{equation}
A^{\mathrm{VICT}}_{i,t}
=\sum_{j\in C_i^\star}\hat{\delta}_{i,j}\omega_{i,t,j}/Z_{i,j}
\label{eq:proof_weighted_adv}
\end{equation}
Here $\omega_{i,t,j}=\sum_{r:(t,j,r)\in E_i}\beta_r$, where $\beta_r$ is a fixed nonnegative relation weight set before training; the default uses $\beta_r=1$ for direct verifier witnesses and splits evidence-path credit uniformly over the support path. Thus $\omega_{i,t,j}=0$ when no witness edge links $(t,j)$, and the denominator bounds each atom's total contribution.

The correction is eligible by construction in the following verifier-backed sense:
\begin{equation}
A^{\mathrm{VICT}}_{i,t}\neq 0
\Rightarrow \exists j,r:\ \mathrm{Gate}_{i,t,j,r}=1 .
\label{eq:proof_carrying_invariant}
\end{equation}
\paragraph{Proposition 1 (Eligibility soundness).}
For any task instance $x$, rollout $\tau_i$, and action $a_{i,t}$, if $A^{\mathrm{VICT}}_{i,t}\neq0$, then there exists at least one atom $j$ and witness relation $r$ such that: (i) the verifier interface reconstructs the terminal verifier within tolerance; (ii) $j$ belongs to a dependency-valid core $C_i^\star$; (iii) $(t,j,r)$ is an observed proof edge; (iv) the atom marginal is non-zero; and (v) the proof record satisfies the dependency constraints in $D_x$.
This proposition does not claim causal necessity or optimality of the action; it guarantees only verifier-backed eligibility. Appendix~\ref{app:eligibility_proof} gives the proof sketch.

The final action advantage keeps the outcome signal as the anchor and adds only a clipped verifier-derived correction:
\begin{equation}
A_{i,t}^{\mathrm{final}}=A_i^{\mathrm{base}}+\lambda\,\mathrm{clip}(A^{\mathrm{VICT}}_{i,t},-c,c)
\label{eq:final_adv}
\end{equation}
The same interface composes with a rollout-side method $M$ when it exposes an action-aligned advantage $A^M_{i,t}$ and a separate rollout-level outcome component $A_i^{\mathrm{out}}$. We set $q_i=\operatorname{sign}_{\epsilon}(A_i^{\mathrm{out}})$ and use
\begin{equation}
A_{i,t}^{M+\mathrm{VICT}}=A^M_{i,t}+\lambda_M\,\mathrm{clip}(A^{\mathrm{VICT}}_{i,t},-c_M,c_M),
\label{eq:hybrid_adv}
\end{equation}
where $\lambda_M$ and $c_M$ are calibrated from $A^M$ before adding \textsc{VICT}. This calibrates numerical scale without assuming that the two credit signals are statistically independent.
\begin{table*}[!t]
    \vspace{-0.4em}
    \small
    \centering
    \setlength{\tabcolsep}{5.2pt}
    \renewcommand{\arraystretch}{0.98}
    \caption{Main performance on ALFWorld and WebShop. ALFWorld reports task-wise and average success rate (\%), while WebShop reports normalized score and strict success rate (\%). Subscripts denote standard deviations where available.}
    \label{tab:main_alfworld_webshop}
    \resizebox{\textwidth}{!}{
    \begin{tabular}{lccccccccc}
        \toprule
        Method &
        \multicolumn{7}{c}{ALFWorld} &
        \multicolumn{2}{c}{WebShop} \\
        \cmidrule(lr){2-8}\cmidrule(lr){9-10}
        & Pick & Look & Clean & Heat & Cool & Pick2 & All & Score & Succ. \\
        \midrule
        \multicolumn{10}{c}{Closed-Source LLMs} \\
        GPT-4o & 75.3 & 60.8 & 31.2 & 56.7 & 21.6 & 49.8 & 48.0 & 31.8 & 23.7 \\
        Gemini-2.5-Pro & 92.8 & 63.3 & 62.1 & 69.0 & 26.6 & 58.7 & 60.3 & 42.5 & 35.9 \\
        \midrule
        \multicolumn{10}{c}{Qwen2.5-1.5B-Instruct} \\
        Base & 5.9 & 5.5 & 3.3 & 9.7 & 4.2 & 0.0 & 4.1 & 23.1 & 5.2 \\
        ReAct & 17.4 & 20.5 & 15.7 & 6.2 & 7.7 & 2.0 & 12.8 & 40.1 & 11.3 \\
        Reflexion & 35.3 & 22.2 & 21.7 & 13.6 & 19.4 & 3.7 & 21.8 & 55.8 & 21.9 \\
        RLOO & 88.3\textsubscript{\textpm3.0} & 52.8\textsubscript{\textpm8.6} & 71.0\textsubscript{\textpm5.9} & 62.8\textsubscript{\textpm8.7} & 66.4\textsubscript{\textpm5.5} & 56.9\textsubscript{\textpm4.7} & 69.7\textsubscript{\textpm2.5} & 73.9\textsubscript{\textpm5.6} & 52.1\textsubscript{\textpm6.7} \\
        GRPO & 85.3\textsubscript{\textpm1.5} & 53.7\textsubscript{\textpm8.0} & 84.5\textsubscript{\textpm6.8} & 78.2\textsubscript{\textpm7.9} & 59.7\textsubscript{\textpm5.0} & 53.5\textsubscript{\textpm5.6} & 72.8\textsubscript{\textpm3.6} & 75.8\textsubscript{\textpm3.5} & 56.8\textsubscript{\textpm3.8} \\
        GiGPO & 94.4\textsubscript{\textpm5.9} & 67.5\textsubscript{\textpm4.6} & 94.8\textsubscript{\textpm3.8} & 94.4\textsubscript{\textpm7.8} & 79.8\textsubscript{\textpm4.7} & 76.4\textsubscript{\textpm5.4} & 86.7\textsubscript{\textpm1.7} & 83.1\textsubscript{\textpm1.6} & 65.0\textsubscript{\textpm3.2} \\
        SALT & 96.2\textsubscript{\textpm1.7} & 65.2\textsubscript{\textpm10.8} & 93.1\textsubscript{\textpm4.7} & 81.8\textsubscript{\textpm8.3} & 85.0\textsubscript{\textpm6.9} & 77.0\textsubscript{\textpm4.7} & 85.2\textsubscript{\textpm2.5} & 86.9\textsubscript{\textpm0.6} & 74.7\textsubscript{\textpm2.4} \\
        HCAPO & 88.6\textsubscript{\textpm7.0} & 75.0\textsubscript{\textpm0.0} & 97.6\textsubscript{\textpm1.8} & 90.7\textsubscript{\textpm6.9} & 84.2\textsubscript{\textpm0.0} & 74.2\textsubscript{\textpm6.9} & 87.0\textsubscript{\textpm4.1} & 83.8\textsubscript{\textpm0.7} & 68.5\textsubscript{\textpm1.0} \\
        \rowcolor{blue!8}
        \textsc{VICT} (ours) & 95.2\textsubscript{\textpm1.0} & \textbf{84.0}\textsubscript{\textpm2.2} & 96.0\textsubscript{\textpm1.4} & 92.0\textsubscript{\textpm2.4} & \textbf{89.0}\textsubscript{\textpm2.0} & \textbf{89.8}\textsubscript{\textpm2.4} & \textbf{91.0}\textsubscript{\textpm0.9} & \textbf{90.6}\textsubscript{\textpm0.7} & \textbf{81.7}\textsubscript{\textpm1.1} \\
        \midrule
        \multicolumn{10}{c}{Qwen2.5-7B-Instruct} \\
        Base & 33.4 & 21.6 & 19.3 & 6.9 & 2.8 & 3.2 & 14.8 & 26.4 & 7.8 \\
        ReAct & 48.5 & 35.4 & 34.3 & 13.2 & 18.2 & 17.6 & 31.2 & 46.2 & 19.5 \\
        Reflexion & 62.0 & 41.6 & 44.9 & 30.9 & 36.3 & 23.8 & 42.7 & 58.1 & 28.8 \\
        RLOO & 87.6\textsubscript{\textpm4.3} & 78.2\textsubscript{\textpm8.3} & 87.3\textsubscript{\textpm5.8} & 81.3\textsubscript{\textpm7.6} & 71.9\textsubscript{\textpm5.2} & 48.9\textsubscript{\textpm8.4} & 75.5\textsubscript{\textpm4.6} & 80.3\textsubscript{\textpm3.2} & 65.7\textsubscript{\textpm4.0} \\
        GRPO & 90.8\textsubscript{\textpm5.1} & 66.1\textsubscript{\textpm6.7} & 89.3\textsubscript{\textpm5.4} & 74.7\textsubscript{\textpm6.9} & 72.5\textsubscript{\textpm5.4} & 64.7\textsubscript{\textpm7.3} & 77.6\textsubscript{\textpm5.2} & 79.3\textsubscript{\textpm2.8} & 66.1\textsubscript{\textpm3.7} \\
        GiGPO & 97.7\textsubscript{\textpm1.6} & 82.7\textsubscript{\textpm7.9} & 98.8\textsubscript{\textpm1.6} & 83.7\textsubscript{\textpm7.2} & 89.3\textsubscript{\textpm8.2} & 79.2\textsubscript{\textpm6.6} & 90.8\textsubscript{\textpm1.3} & 84.4\textsubscript{\textpm2.9} & 72.8\textsubscript{\textpm3.2} \\
        SALT & 87.5\textsubscript{\textpm4.9} & 58.8\textsubscript{\textpm14.7} & 89.3\textsubscript{\textpm3.6} & 75.3\textsubscript{\textpm4.3} & 70.2\textsubscript{\textpm5.5} & 68.5\textsubscript{\textpm5.8} & 77.8\textsubscript{\textpm1.7} & 84.7\textsubscript{\textpm1.7} & 76.2\textsubscript{\textpm3.4} \\
        HCAPO & 99.1\textsubscript{\textpm1.3} & 90.3\textsubscript{\textpm2.0} & 97.3\textsubscript{\textpm1.9} & 81.8\textsubscript{\textpm8.8} & 90.8\textsubscript{\textpm6.6} & 81.9\textsubscript{\textpm10.0} & 91.4\textsubscript{\textpm2.3} & 85.1\textsubscript{\textpm1.3} & 73.8\textsubscript{\textpm2.8} \\
        \rowcolor{blue!8}
        \textsc{VICT} (ours) & \textbf{99.2}\textsubscript{\textpm0.6} & \textbf{91.5}\textsubscript{\textpm1.8} & 98.5\textsubscript{\textpm1.0} & \textbf{92.2}\textsubscript{\textpm2.0} & \textbf{93.8}\textsubscript{\textpm1.6} & \textbf{86.8}\textsubscript{\textpm2.4} & \textbf{93.7}\textsubscript{\textpm0.8} & \textbf{91.2}\textsubscript{\textpm0.6} & \textbf{83.6}\textsubscript{\textpm0.9} \\
        \bottomrule
    \end{tabular}
    }
\end{table*}

This advantage is plugged into the standard clipped policy-gradient objective with KL regularization. \textsc{VICT} changes only the advantage tensor: it trains no critic, requires no process labels, adds no branch rollouts, and exposes no verifier atoms at inference time.
Every non-zero correction is logged with its action, atom, witness, evidence source, core marginal, witness weight, and assigned correction. The reward-shaping scope is discussed in Appendix~\ref{app:policy_invariance_scope}.

\section{Experiments}
\label{sec:experiments}

\subsection{Experimental Setup}
\label{sec:exp_setup}

\paragraph{Benchmarks and metrics.}
We evaluate on three programmatically verifiable agent benchmarks. \textbf{ALFWorld} tests text-based household tasks and reports success across six task types and their average~\citep{shridhar2020alfworld}. \textbf{WebShop} tests product search and purchase under category, attribute, option, and budget constraints, reporting normalized score and strict success~\citep{yao2022webshop}. \textbf{$\tau$-bench} tests service-domain tool-agent-user interaction, reporting Retail/Airline pass@1~\citep{yao2024tau}.

\paragraph{Baselines.}
For ALFWorld and WebShop, we compare against closed-source LLMs, Qwen2.5-Instruct prompting, ReAct~\citep{yao2023react}, Reflexion~\citep{shinn2023reflexion}, outcome-level RLOO/GRPO~\citep{ahmadian2024back,shao2024deepseekmath}, and fine-grained credit baselines GiGPO, SALT, and HCAPO~\citep{feng2025group,li2026salt,tan2026hcapo}. For $\tau$-bench, we compare Qwen3-8B variants from the Fission-GRPO protocol: Base, GRPO, DAPO, Dr.GRPO, AWPO, and Fission-GRPO~\citep{yu2025dapo,liu2025understanding,lin2025awpo,zhang2026fissiongrpo}.

\paragraph{Training and instrumentation.}
For \textsc{VICT} on ALFWorld and WebShop, we use Qwen2.5-1.5B/7B-Instruct, group size $N=8$, learning rate $1\times10^{-6}$, KL coefficient $0.01$, and 50/15-step caps; \textsc{VICT} rows report mean and standard deviation over three seeds. On $\tau$-bench, Qwen3-8B pass@1 uses simulated-user interaction. \textsc{VICT} changes only training-time advantages and exposes no verifier atoms at inference. Its adapters require 118--238 LoC and 6.5--11.5 hours, with 11.9--16.7\% training-time overhead on the two primary benchmarks (Appendix Figure~\ref{fig:cost_overhead}).

\subsection{Main Results on ALFWorld and WebShop}
\label{sec:main_alfworld_webshop}

\begin{figure}[t]
    \centering
    \includegraphics[width=\columnwidth]{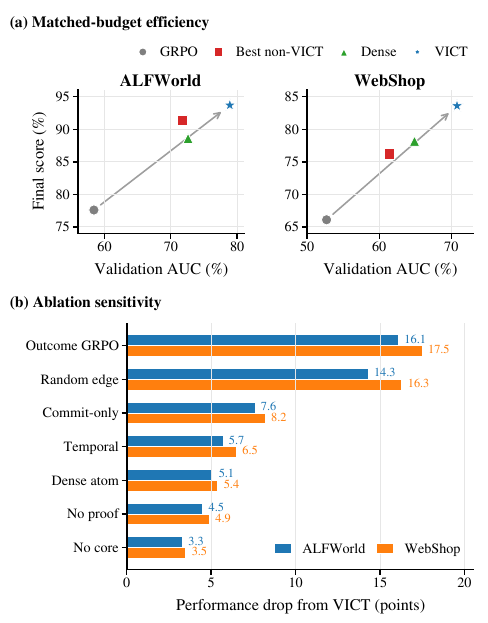}
    \caption{Core evidence with Qwen2.5-7B. (a) Final score versus normalized validation AUC over the same 300 updates; arrows connect outcome-only GRPO to \textsc{VICT}, and upper right is better. ``Best non-\textsc{VICT}'' is HCAPO for ALFWorld and SALT for WebShop. (b) Point decrease from full \textsc{VICT} for each ablation or negative control. Exact mean and standard deviation values are in Table~\ref{tab:ablation} and Appendix Table~\ref{tab:sample_efficiency}.}
    \label{fig:core_evidence}
    \vspace{-0.5em}
\end{figure}

Table~\ref{tab:main_alfworld_webshop} summarizes performance on ALFWorld and WebShop. With Qwen2.5-1.5B, \textsc{VICT} improves over GRPO by 18.2 points on ALFWorld average success and 24.9 points on WebShop strict success. With Qwen2.5-7B, where several ALFWorld subtasks are near saturation, \textsc{VICT} reaches 93.7 average success and 83.6 WebShop strict success, corresponding to gains of 16.1 and 17.5 points over GRPO.
The largest gains appear on task types where the verifier exposes delayed or commit-sensitive facts: \textit{Look}, \textit{Cool}, and \textit{Pick2} in ALFWorld, and final purchase correctness in WebShop. This pattern supports the main hypothesis: when a failed trajectory contains useful search or state-changing actions before a wrong commit, verifier-backed credit can reward the supported actions without reinforcing the final mistake. Figure~\ref{fig:core_evidence}(a) further shows that the final gains coincide with higher validation AUC over the same 300-update budget; Appendix~\ref{app:sample_efficiency} reports the exact values.

The task-wise pattern also matches verifier-backed credit: ALFWorld's 7B setting is near saturation on \textit{Pick}, \textit{Clean}, and \textit{Heat}, while improvements concentrate on \textit{Look}, \textit{Cool}, and \textit{Pick2}, where observations or transformations are easy to overwrite. WebShop is less saturated because strict success penalizes one wrong hard attribute or premature purchase, so separating evidence-gathering from commits yields a larger gain. This keeps the claim tied to training-time credit assignment rather than policy access to more inference-time information.

Relative to the strongest fine-grained baseline in each primary block, the margin is smaller than against GRPO but remains positive at both model scales. With Qwen2.5-1.5B, \textsc{VICT} is 4.0 points above HCAPO on ALFWorld average and 7.0 points above SALT on WebShop strict success. With Qwen2.5-7B, the corresponding margins are 2.3 and 7.4 points. The narrower ALFWorld margin at 7B is consistent with the benchmark nearing saturation, whereas WebShop retains room for errors in product identity and hard attributes. We therefore read the primary result as consistent improvement across scale and benchmark, not as evidence that one credit rule dominates every task or subcategory.

\begin{table}[t]
    \centering
    \small
    \setlength{\tabcolsep}{3.6pt}
    \renewcommand{\arraystretch}{1.02}
    \caption{Compatibility with rollout-side credit methods using Qwen2.5-7B. Each cell reports the method alone / with \textsc{VICT}; comparisons are descriptive.}
    \label{tab:hybrid_credit}
    \resizebox{\columnwidth}{!}{
    \begin{tabular}{lcc}
        \toprule
        Setting & ALFWorld All & WebShop strict \\
        \midrule
        \textsc{VICT} alone & 93.7\textsubscript{\textpm0.8} & 83.6\textsubscript{\textpm0.9} \\
        GiGPO alone / +\textsc{VICT} & 90.8\textsubscript{\textpm1.3} / \textbf{94.6}\textsubscript{\textpm0.7} & 72.8\textsubscript{\textpm3.2} / 84.7\textsubscript{\textpm1.1} \\
        HCAPO alone / +\textsc{VICT} & 91.4\textsubscript{\textpm2.3} / 94.3\textsubscript{\textpm0.9} & -- \\
        SALT alone / +\textsc{VICT} & -- & 76.2\textsubscript{\textpm3.4} / \textbf{85.2}\textsubscript{\textpm0.8} \\
        \bottomrule
    \end{tabular}
    }
    \vspace{-0.5em}
\end{table}

\paragraph{Compatibility with rollout-side credit.}
Table~\ref{tab:hybrid_credit} evaluates Eq.~\ref{eq:hybrid_adv}. Each combination is above both \textsc{VICT} alone and the corresponding rollout-side method alone. We treat this result as descriptive and scope compatibility to methods that expose both an action-aligned advantage and an independent rollout-level outcome component.

The increments over the standalone rollout-side methods are 3.8/11.9 points for GiGPO on ALFWorld/WebShop, 2.9 points for HCAPO on ALFWorld, and 9.0 points for SALT on WebShop. The gains over \textsc{VICT} alone are smaller, ranging from 0.6 to 1.6 points. This asymmetry is consistent with the two signals sharing some useful trajectory information while correcting different errors. Because the table does not contain every method--domain pair or a factorial interaction analysis, it supports practical compatibility rather than a claim of universal additivity.

\subsection{Additional Evidence on $\tau$-Bench}
\label{sec:tau_bench}

The service-agent setting tests a different verifier structure: reward depends on satisfying user requests while preserving backend state and domain policies. Under the Fission-GRPO $\tau$-bench protocol, \textsc{VICT} reaches 56.6/45.1 Retail/Airline pass@1, compared with 51.3/40.0 for Fission-GRPO; Table~\ref{tab:tau_bench} gives the full comparison. Because this comparison uses a different backbone and protocol from the primary experiments, we treat it as supplemental evidence. The pattern is consistent with verifier tracing: database edits, confirmations, forbidden updates, and final-response checks create atoms whose witnesses localize to API calls. The Airline degradation of DAPO relative to Base should not be over-interpreted.

The improvement over Fission-GRPO is similar in Retail and Airline, at 5.3 and 5.1 points. This agreement is useful as a cross-domain check of the verifier interface because the two domains emphasize different API and policy constraints. It should not be read as a direct comparison with the Qwen2.5 experiments: the backbone, simulated-user setting, and protocol all differ, so only the within-table comparisons are informative.

\begin{table}[!t]
    \centering
    \small
    \setlength{\tabcolsep}{8pt}
    \renewcommand{\arraystretch}{1.05}
    \caption{Supplemental service-agent evidence on $\tau$-bench. Pass@1 (\%) is measured with Qwen3-8B under simulated-user interaction; the \textsc{VICT} row reports mean and standard deviation over three seeds.}
    \label{tab:tau_bench}
    \begin{tabular}{lcc}
        \toprule
        Method & Retail & Airline \\
        \midrule
        Base & 35.7 & 23.2 \\
        GRPO & 39.1 & 36.0 \\
        DAPO & 45.2 & 20.0 \\
        Dr.GRPO & 47.0 & 28.0 \\
        AWPO & 43.5 & 30.0 \\
        Fission-GRPO & 51.3 & 40.0 \\
        \midrule
        \textsc{VICT} (ours) & \textbf{56.6}\textsubscript{\textpm0.8} & \textbf{45.1}\textsubscript{\textpm1.0} \\
        \bottomrule
    \end{tabular}
\end{table}

\subsection{Ablations and Diagnostics}
\label{sec:analysis}

Figure~\ref{fig:core_evidence}(b) isolates the choices that distinguish \textsc{VICT} from dense verifier rewards and heuristic step credit. ALFWorld and WebShop use Qwen2.5-7B-Instruct to match the 7B block in Table~\ref{tab:main_alfworld_webshop}; $\tau$-bench is excluded because it uses a different backbone and protocol. Every alternative reduces both primary metrics. Outcome-only GRPO and randomized proof edges produce the largest drops, while dense atom rewards, commit-only credit, and temporal-nearest credit show that simply exposing verifier facts, crediting final actions, or using temporal proximity is insufficient. Removing the dependency core or proof edges also costs 3.3--4.9 points. These controls reuse the same adapters, atoms, and evidence extractors and change only the credit rule. Their consistent degradation suggests that the gain comes from combining verifier relevance with observed evidence and group-normalized correction; Table~\ref{tab:ablation} reports the exact three-seed values.

The controls also form a coherent severity ordering. Outcome-only GRPO and randomized proof edges lose 14.3--17.5 points, indicating that both localized verifier information and correct action--atom alignment matter. Commit-only, temporal-nearest, and dense-atom variants lose 5.1--8.2 points, while removing only the dependency core or proof edges loses 3.3--4.9 points. These values do not isolate statistically independent causal effects because the components interact, but they argue against the simpler interpretation that any dense verifier-derived signal or any sparse final-action signal is sufficient.

\begin{table}[!t]
    \centering
    \small
    \setlength{\tabcolsep}{4.0pt}
    \renewcommand{\arraystretch}{0.98}
    \caption{Ablation and negative-control results with Qwen2.5-7B. Values are mean with standard deviation over three seeds.}
    \label{tab:ablation}
    \resizebox{\columnwidth}{!}{
    \begin{tabular}{lcc}
        \toprule
        Variant & ALFWorld & WebShop \\
        \midrule
        Outcome-only GRPO & 77.6\textsubscript{\textpm5.2} & 66.1\textsubscript{\textpm3.7} \\
        Dense atom reward & 88.6\textsubscript{\textpm1.7} & 78.2\textsubscript{\textpm1.5} \\
        Commit-only verifier credit & 86.1\textsubscript{\textpm2.2} & 75.4\textsubscript{\textpm1.8} \\
        Temporal-nearest atom credit & 88.0\textsubscript{\textpm2.0} & 77.1\textsubscript{\textpm1.7} \\
        Randomized proof-edge placebo & 79.4\textsubscript{\textpm3.9} & 67.3\textsubscript{\textpm3.4} \\
        w/o dependency core & 90.4\textsubscript{\textpm1.5} & 80.1\textsubscript{\textpm1.2} \\
        w/o proof edges & 89.2\textsubscript{\textpm1.8} & 78.7\textsubscript{\textpm1.6} \\
        \textsc{VICT} (ours) & \textbf{93.7}\textsubscript{\textpm0.8} & \textbf{83.6}\textsubscript{\textpm0.9} \\
        \bottomrule
    \end{tabular}
    }
\end{table}

\begin{table}[t]
    \centering
    \small
    \setlength{\tabcolsep}{4.5pt}
    \renewcommand{\arraystretch}{0.96}
    \caption{Verifier-interface diagnostics before policy updates. Reconstruction and mutation conformance test agreement with the original verifier; eligibility-invariant pass rate, coverage, and abstention audit whether verifier facts can be safely attached to actions.}
    \label{tab:verifier_diagnostics}
    \resizebox{\columnwidth}{!}{
    \begin{tabular}{lcccccc}
        \toprule
        Domain & Atoms & Recon. & Mutation & Elig. pass & Proof cov. & Abstain \\
        \midrule
        ALFWorld & 5.8 & 100.0 & 99.6 & 100.0 & 92.4 & 8.1 \\
        WebShop & 6.3 & 99.8 & 98.9 & 100.0 & 88.7 & 11.6 \\
        $\tau$ Retail & 9.7 & 100.0 & 99.2 & 100.0 & 86.5 & 14.3 \\
        $\tau$ Airline & 11.2 & 99.7 & 98.6 & 100.0 & 84.9 & 15.8 \\
        \bottomrule
    \end{tabular}
    }
    \vspace{-0.7em}
\end{table}

We audit the verifier interface before policy updates. Table~\ref{tab:verifier_diagnostics} reports high reward reconstruction and mutation conformance, indicating that the atom aggregator matches the original terminal verifier rather than an auxiliary rubric. The eligibility-invariant pass rate is 100\% because non-zero corrections are emitted only after Eq.~\ref{eq:proof_carrying_invariant} passes; lower values would indicate an implementation error, not causal accuracy. Proof coverage is lower than atom coverage because \textsc{VICT} abstains on terminal-only facts without a reliable writer or commit action. This abstention is intentional: unsupported atoms should not become action-level gradients. We therefore read the diagnostic table as a safety gate rather than a performance proxy: high conformance makes the credit interface admissible, while coverage and abstention indicate how often the method can use that interface without guessing about unsupported actions across domains and training stages, not just whether implementations satisfy invariants. Appendices~\ref{app:verifier_interface_construction}--\ref{app:diagnostics} report credit sparsity, trigger rate, core size, budget-hit rate, and runtime overhead, which better characterize training behavior than the invariant pass rate alone. The average atom counts remain modest, keeping the interface tractable.

The domain pattern provides an additional check on this interpretation. ALFWorld/WebShop have 5.8/6.3 atoms on average, proof coverage of 92.4/88.7\%, and abstention of 8.1/11.6\%; $\tau$ Retail/Airline have 9.7/11.2 atoms, coverage of 86.5/84.9\%, and abstention of 14.3/15.8\%. Thus, the more structured service domains expose more atoms but also leave more facts without safe action witnesses. \textsc{VICT} responds by abstaining more often rather than forcing denser credit, which is the intended behavior of the proof gate. These are interface diagnostics, not independent evidence that the attributed actions are causally necessary.

\paragraph{Qualitative behavior.}
Across the three benchmarks, \textsc{VICT} changes the learned behavior in the places predicted by the proof graph. In WebShop, agents delay purchase until hard attributes and options are observed, rather than buying a semantically close product after the first relevant search result. In ALFWorld, agents more reliably preserve intermediate achievements such as acquiring the correct object before transformation and placement. In $\tau$-bench, agents learn to request missing confirmation and avoid irrelevant database writes before committing API updates. These behaviors align with the actions that receive verifier-backed credit: evidence-revealing actions, direct state-changing actions, and commit actions whose verifier atoms are satisfied at the moment of commitment.

\section{Conclusion}
\label{sec:conclusion}

This paper introduced \textsc{VICT}, a verifier-instrumented credit tracing method for long-horizon LLM agent reinforcement learning with instrumentable programmatic verifiers. \textsc{VICT} exposes verifier atoms, dependencies, evidence maps, and commit predicates, then assigns sparse action-level advantage corrections only through dependency-valid proof edges. It preserves the original outcome reward, needs no critic or process labels, and keeps non-zero corrections auditable. Across embodied, web, and tool-use settings, experiments and diagnostics show that verifier-backed credit improves fine-grained training while satisfying reconstruction, mutation, eligibility-invariant, sparsity, and abstention checks.

\section*{Limitations}

\textsc{VICT} is limited to settings where terminal rewards come from verifiers that can be exposed as executable or evidence-backed atoms, and where relevant state changes, evidence reveals, commits, or violations are observable in trajectory logs. It is less direct for holistic learned judges, hidden verifier state, or tasks dominated by exploration failure. Extending it to LLM-as-a-judge settings would require auditable, evidence-grounded atom construction rather than prompt-generated criteria alone.

Verifier instrumentation also has a real engineering cost. Level-0 state-diff atoms are often cheap, but Level-1 branch instrumentation and Level-2 deterministic adapters may require domain knowledge and maintenance as verifiers evolve. Each deployment should therefore report adapter lines of code, person-hours, atom coverage, and the Level-0/1/2 atom mix.

\textsc{VICT} further depends on small, dependency-closed verifier cores. Large atom sets or ambiguous dependencies can make greedy search return larger local cores or abstain, preserving the proof-edge invariant but reducing credit recall. Proof edges certify verifier-defined eligibility rather than causal necessity: redundant or correlated support can receive credit, and witness-mapping errors can survive conformance tests. Finally, because \textsc{VICT} only changes the training-time advantage tensor, it is not a policy-invariant shaping guarantee and still requires strong exploration, stable optimization, careful verifier design, and matched-budget diagnostics.

\bibliography{ref}

@inproceedings{yao2023react,
  title = {{ReAct}: Synergizing Reasoning and Acting in Language Models},
  author = {Yao, Shunyu and Zhao, Jeffrey and Yu, Dian and Du, Nan and Shafran, Izhak and Narasimhan, Karthik and Cao, Yuan},
  booktitle = {International Conference on Learning Representations},
  year = {2023},
  url = {https://openreview.net/forum?id=WE_vluYUL-X}
}

@article{schick2023toolformer,
  title = {{Toolformer}: Language Models Can Teach Themselves to Use Tools},
  author = {Schick, Timo and Dwivedi-Yu, Jane and Dess{\`i}, Roberto and Raileanu, Roberta and Lomeli, Maria and Zettlemoyer, Luke and Cancedda, Nicola and Scialom, Thomas},
  journal = {arXiv preprint arXiv:2302.04761},
  year = {2023},
  url = {https://arxiv.org/abs/2302.04761}
}

@inproceedings{shridhar2020alfworld,
  title = {{ALFWorld}: Aligning Text and Embodied Environments for Interactive Learning},
  author = {Shridhar, Mohit and Yuan, Xingdi and C{\^o}t{\'e}, Marc-Alexandre and Bisk, Yonatan and Trischler, Adam and Hausknecht, Matthew},
  booktitle = {International Conference on Learning Representations},
  year = {2021},
  url = {https://openreview.net/forum?id=0IOX0YcCdTn}
}

@inproceedings{yao2022webshop,
  title = {{WebShop}: Towards Scalable Real-World Web Interaction with Grounded Language Agents},
  author = {Yao, Shunyu and Chen, Howard and Yang, John and Narasimhan, Karthik},
  booktitle = {Advances in Neural Information Processing Systems},
  year = {2022}
}

@inproceedings{ahmadian2024back,
  title = {Back to Basics: Revisiting {REINFORCE} Style Optimization for Learning from Human Feedback in {LLMs}},
  author = {Ahmadian, Arash and Cremer, Chris and Gall{\'e}, Matthias and Fadaee, Marzieh and Kreutzer, Julia and Pietquin, Olivier and {\"U}st{\"u}n, Ahmet and Hooker, Sara},
  booktitle = {Proceedings of the 62nd Annual Meeting of the Association for Computational Linguistics (Volume 1: Long Papers)},
  pages = {12248--12267},
  year = {2024}
}

@article{shao2024deepseekmath,
  title = {{DeepSeekMath}: Pushing the Limits of Mathematical Reasoning in Open Language Models},
  author = {Shao, Zhihong and Wang, Peiyi and Zhu, Qihao and Xu, Runxin and Song, Junxiao and Bi, Xiao and Zhang, Haowei and Zhang, Mingchuan and Li, Y. K. and Wu, Y. and others},
  journal = {arXiv preprint arXiv:2402.03300},
  year = {2024},
  url = {https://arxiv.org/abs/2402.03300}
}

@inproceedings{zeng2024agenttuning,
  title = {{AgentTuning}: Enabling Generalized Agent Abilities for {LLMs}},
  author = {Zeng, Aohan and Liu, Mingdao and Lu, Rui and Wang, Bowen and Liu, Xiao and Dong, Yuxiao and Tang, Jie},
  booktitle = {Findings of the Association for Computational Linguistics: ACL 2024},
  pages = {3053--3077},
  year = {2024},
  address = {Bangkok, Thailand},
  publisher = {Association for Computational Linguistics},
  doi = {10.18653/v1/2024.findings-acl.181},
  url = {https://aclanthology.org/2024.findings-acl.181/}
}

@inproceedings{trivedi2024appworld,
  title = {{AppWorld}: A Controllable World of Apps and People for Benchmarking Interactive Coding Agents},
  author = {Trivedi, Harsh and Khot, Tushar and Hartmann, Mareike and Manku, Ruskin and Dong, Vinty and Li, Edward and Gupta, Shashank and Sabharwal, Ashish and Balasubramanian, Niranjan},
  booktitle = {Proceedings of the 62nd Annual Meeting of the Association for Computational Linguistics (Volume 1: Long Papers)},
  pages = {16022--16076},
  year = {2024}
}

@article{schulman2017proximal,
  title = {Proximal Policy Optimization Algorithms},
  author = {Schulman, John and Wolski, Filip and Dhariwal, Prafulla and Radford, Alec and Klimov, Oleg},
  journal = {arXiv preprint arXiv:1707.06347},
  year = {2017},
  url = {https://arxiv.org/abs/1707.06347}
}

@article{jin2025search,
  title = {Search-{R1}: Training {LLMs} to Reason and Leverage Search Engines with Reinforcement Learning},
  author = {Jin, Bowen and Zeng, Hansi and Yue, Zhenrui and Yoon, Jinsung and Arik, Sercan and Wang, Dong and Zamani, Hamed and Han, Jiawei},
  journal = {arXiv preprint arXiv:2503.09516},
  year = {2025},
  url = {https://arxiv.org/abs/2503.09516}
}

@article{yao2024tau,
  title = {{$\tau$-Bench}: A Benchmark for Tool-Agent-User Interaction in Real-World Domains},
  author = {Yao, Shunyu and Shinn, Noah and Razavi, Pedram and Narasimhan, Karthik},
  journal = {arXiv preprint arXiv:2406.12045},
  year = {2024},
  url = {https://arxiv.org/abs/2406.12045}
}

@article{feng2025group,
  title = {Group-in-Group Policy Optimization for {LLM} Agent Training},
  author = {Feng, Lang and Xue, Zhenghai and Liu, Tingcong and An, Bo},
  journal = {arXiv preprint arXiv:2505.10978},
  year = {2025},
  url = {https://arxiv.org/abs/2505.10978}
}

@inproceedings{li2026salt,
  title = {{SALT}: Step-level Advantage Assignment for Long-horizon Agents via Trajectory Graph},
  author = {Li, Jiazheng and Wang, Yawei and Yan, Qiaojing and Tian, Yijun and Xu, Zhichao and Song, Huan and Xu, Panpan and Cheong, Lin Lee},
  booktitle = {Findings of the Association for Computational Linguistics: EACL 2026},
  pages = {4709--4725},
  year = {2026},
  address = {Rabat, Morocco},
  publisher = {Association for Computational Linguistics},
  doi = {10.18653/v1/2026.findings-eacl.247},
  url = {https://aclanthology.org/2026.findings-eacl.247/}
}

@misc{fang2026proxmo,
  title = {Proximity-Based Multi-Turn Optimization: Practical Credit Assignment for {LLM} Agent Training},
  author = {Fang, Yangyi and Lin, Jiaye and Fu, Xiaoliang and Qin, Cong and Shi, Haolin and Liu, Chang and Zhao, Peilin},
  year = {2026},
  eprint = {2602.19225},
  archivePrefix = {arXiv},
  primaryClass = {cs.AI},
  url = {https://arxiv.org/abs/2602.19225}
}

@misc{tan2026hcapo,
  title = {Hindsight Credit Assignment for Long-Horizon {LLM} Agents},
  author = {Tan, Hui-Ze and Yang, Xiao-Wen and Chen, Hao and Shao, Jie-Jing and Wen, Yi and Shen, Yuteng and Luo, Weihong and Du, Xiku and Guo, Lan-Zhe and Li, Yu-Feng},
  year = {2026},
  eprint = {2603.08754},
  archivePrefix = {arXiv},
  primaryClass = {cs.LG},
  url = {https://arxiv.org/abs/2603.08754}
}

@misc{peng2026hiper,
  title = {{HiPER}: Hierarchical Reinforcement Learning with Explicit Credit Assignment for Large Language Model Agents},
  author = {Peng, Jiangweizhi and Liu, Yuanxin and Zhou, Ruida and Fleming, Charles and Wang, Zhaoran and Garcia, Alfredo and Hong, Mingyi},
  year = {2026},
  eprint = {2602.16165},
  archivePrefix = {arXiv},
  primaryClass = {cs.LG},
  url = {https://arxiv.org/abs/2602.16165}
}

@misc{wang2026subgoal,
  title = {A Subgoal-driven Framework for Improving Long-Horizon {LLM} Agents},
  author = {Wang, Taiyi and Gooding, Sian and Hartmann, Florian and Riva, Oriana and Grefenstette, Edward},
  year = {2026},
  eprint = {2603.19685},
  archivePrefix = {arXiv},
  primaryClass = {cs.AI},
  url = {https://arxiv.org/abs/2603.19685}
}

@misc{xue2026strata,
  title = {{StraTA}: Incentivizing Agentic Reinforcement Learning with Strategic Trajectory Abstraction},
  author = {Xue, Xiangyuan and Zhou, Yifan and Wang, Zidong and Tang, Shengji and Torr, Philip and Ouyang, Wanli and Bai, Lei and Yin, Zhenfei},
  year = {2026},
  eprint = {2605.06642},
  archivePrefix = {arXiv},
  primaryClass = {cs.CL},
  url = {https://arxiv.org/abs/2605.06642}
}

@misc{ji2026treegrpo,
  title = {Tree Search for {LLM} Agent Reinforcement Learning},
  author = {Ji, Yuxiang and Ma, Ziyu and Wang, Yong and Chen, Guanhua and Chu, Xiangxiang and Wu, Liaoni},
  year = {2026},
  eprint = {2509.21240},
  archivePrefix = {arXiv},
  primaryClass = {cs.LG},
  url = {https://arxiv.org/abs/2509.21240}
}

@misc{dong2025arpo,
  title = {Agentic Reinforced Policy Optimization},
  author = {Dong, Guanting and Mao, Hangyu and Ma, Kai and Bao, Licheng and Chen, Yifei and Wang, Zhongyuan and Chen, Zhongxia and Du, Jiazhen and Wang, Huiyang and Zhang, Fuzheng and Zhou, Guorui and Zhu, Yutao and Wen, Ji-Rong and Dou, Zhicheng},
  year = {2025},
  eprint = {2507.19849},
  archivePrefix = {arXiv},
  primaryClass = {cs.LG},
  url = {https://arxiv.org/abs/2507.19849}
}

@misc{zhao2026branpo,
  title = {{BranPO}: Scalable Contrastive Branch Sampling for Long-Horizon Agentic Reinforcement Learning},
  author = {Zhao, Yubao and Huang, Weiquan and Wang, Sudong and Zhao, Ruochen and Chen, Chen and Shu, Yao and Qin, Chengwei},
  year = {2026},
  eprint = {2602.03719},
  archivePrefix = {arXiv},
  primaryClass = {cs.CL},
  url = {https://arxiv.org/abs/2602.03719}
}

@misc{wu2026spark,
  title = {Spark: Strategic Policy-Aware Exploration via Dynamic Branching for Long-Horizon Agentic Learning},
  author = {Wu, Jinyang and Yang, Shuo and Yang, Changpeng and Shen, Yuhao and Zhang, Shuai and Wen, Zhengqi and Tao, Jianhua},
  year = {2026},
  eprint = {2601.20209},
  archivePrefix = {arXiv},
  primaryClass = {cs.LG},
  url = {https://arxiv.org/abs/2601.20209}
}

@inproceedings{uesato2022process,
  title = {Solving Math Word Problems with Process- and Outcome-Based Feedback},
  author = {Uesato, Jonathan and Kushman, Nate and Kumar, Ramana and Song, Francis and Siegel, Noah and Wang, Lisa and Creswell, Antonia and Irving, Geoffrey and Higgins, Irina},
  booktitle = {NeurIPS 2022 Workshop on MATH-AI},
  year = {2022},
  url = {https://arxiv.org/abs/2211.14275}
}

@inproceedings{lightman2023let,
  title = {Let's Verify Step by Step},
  author = {Lightman, Hunter and Kosaraju, Vineet and Burda, Yura and Edwards, Harri and Baker, Bowen and Lee, Teddy and Leike, Jan and Schulman, John and Sutskever, Ilya and Cobbe, Karl},
  booktitle = {International Conference on Learning Representations},
  year = {2024},
  url = {https://arxiv.org/abs/2305.20050}
}

@misc{luo2024omegaprm,
  title = {Improve Mathematical Reasoning in Language Models by Automated Process Supervision},
  author = {Luo, Liangchen and Liu, Yinxiao and Liu, Rosanne and Phatale, Samrat and Guo, Meiqi and Lara, Harsh and Li, Yunxuan and Shu, Lei and Zhu, Yun and Meng, Lei and Sun, Jiao and Rastogi, Abhinav},
  year = {2024},
  eprint = {2406.06592},
  archivePrefix = {arXiv},
  primaryClass = {cs.CL},
  url = {https://arxiv.org/abs/2406.06592}
}

@misc{xi2025agentprm,
  title = {{AgentPRM}: Process Reward Models for {LLM} Agents via Step-Wise Promise and Progress},
  author = {Xi, Zhiheng and Liao, Chenyang and Li, Guanyu and Yang, Yajie and Chen, Wenxiang and Zhang, Zhihao and Wang, Binghai and Jin, Senjie and Zhou, Yuhao and Guan, Jian and Wu, Wei and Ji, Tao and Gui, Tao and Zhang, Qi and Huang, Xuanjing},
  year = {2025},
  eprint = {2511.08325},
  archivePrefix = {arXiv},
  primaryClass = {cs.CL},
  url = {https://arxiv.org/abs/2511.08325}
}

@inproceedings{arjona2019rudder,
  title = {{RUDDER}: Return Decomposition for Delayed Rewards},
  author = {Arjona-Medina, Jose A. and Gillhofer, Michael and Widrich, Michael and Unterthiner, Thomas and Brandstetter, Johannes and Hochreiter, Sepp},
  booktitle = {Advances in Neural Information Processing Systems},
  volume = {32},
  year = {2019},
  url = {https://papers.nips.cc/paper/9509-rudder-return-decomposition-for-delayed-rewards}
}

@inproceedings{ng1999policy,
  title = {Policy Invariance under Reward Transformations: Theory and Application to Reward Shaping},
  author = {Ng, Andrew Y. and Harada, Daishi and Russell, Stuart},
  booktitle = {Proceedings of the Sixteenth International Conference on Machine Learning},
  pages = {278--287},
  year = {1999}
}

@inproceedings{toroicarte2018reward,
  title = {Using Reward Machines for High-Level Task Specification and Decomposition in Reinforcement Learning},
  author = {Icarte, Rodrigo Toro and Klassen, Toryn and Valenzano, Richard and McIlraith, Sheila},
  booktitle = {Proceedings of the 35th International Conference on Machine Learning},
  pages = {2107--2116},
  year = {2018},
  editor = {Dy, Jennifer and Krause, Andreas},
  volume = {80},
  series = {Proceedings of Machine Learning Research},
  publisher = {PMLR},
  url = {https://proceedings.mlr.press/v80/icarte18a.html}
}

@inproceedings{juozapaitis2019explainable,
  title = {Explainable Reinforcement Learning via Reward Decomposition},
  author = {Juozapaitis, Zoe and Koul, Anurag and Fern, Alan and Erwig, Martin and Doshi-Velez, Finale},
  booktitle = {Proceedings of the IJCAI/ECAI Workshop on Explainable Artificial Intelligence},
  year = {2019}
}

@misc{xie2026autorubric,
  title = {Auto-Rubric: Learning From Implicit Weights to Explicit Rubrics for Reward Modeling},
  author = {Xie, Lipeng and Huang, Sen and Zhang, Zhuo and Zou, Anni and Zhai, Yunpeng and Ren, Dingchao and Zhang, Kezun and Hu, Haoyuan and Liu, Boyin and Chen, Haoran and Liu, Zhaoyang and Ding, Bolin},
  year = {2026},
  eprint = {2510.17314},
  archivePrefix = {arXiv},
  primaryClass = {cs.LG},
  url = {https://arxiv.org/abs/2510.17314}
}

@misc{yu2025dece,
  title = {Beyond Pointwise Scores: Decomposed Criteria-Based Evaluation of {LLM} Responses},
  author = {Yu, Fangyi and Seedat, Nabeel and Herrmannova, Dasha and Schilder, Frank and Schwarz, Jonathan Richard},
  year = {2025},
  eprint = {2509.16093},
  archivePrefix = {arXiv},
  primaryClass = {cs.CL},
  url = {https://arxiv.org/abs/2509.16093}
}

@misc{gunjal2025rubrics,
  title = {Rubrics as Rewards: Reinforcement Learning Beyond Verifiable Domains},
  author = {Gunjal, Anisha and Wang, Anthony and Lau, Elaine and Nath, Vaskar and He, Yunzhong and Liu, Bing and Hendryx, Sean},
  year = {2025},
  eprint = {2507.17746},
  archivePrefix = {arXiv},
  primaryClass = {cs.LG},
  url = {https://arxiv.org/abs/2507.17746}
}

@misc{zhang2026atomic,
  title = {Rethinking Atomic Decomposition for {LLM} Judges: A Prompt-Controlled Study of Reference-Grounded {QA} Evaluation},
  author = {Zhang, Xinran},
  year = {2026},
  eprint = {2603.28005},
  archivePrefix = {arXiv},
  primaryClass = {cs.CL},
  url = {https://arxiv.org/abs/2603.28005}
}

@article{shinn2023reflexion,
  title = {Reflexion: Language Agents with Verbal Reinforcement Learning},
  author = {Shinn, Noah and Cassano, Federico and Gopinath, Ashwin and Narasimhan, Karthik and Yao, Shunyu},
  journal = {Advances in Neural Information Processing Systems},
  volume = {36},
  pages = {8634--8652},
  year = {2023}
}

@misc{yu2025dapo,
  title = {{DAPO}: An Open-Source {LLM} Reinforcement Learning System at Scale},
  author = {Yu, Qiying and Zhang, Zheng and Zhu, Ruofei and Yuan, Yufeng and Zuo, Xiaochen and Yue, Yu and Dai, Weinan and Fan, Tiantian and Liu, Gaohong and Liu, Lingjun and others},
  year = {2025},
  eprint = {2503.14476},
  archivePrefix = {arXiv},
  primaryClass = {cs.LG},
  url = {https://arxiv.org/abs/2503.14476}
}

@misc{liu2025understanding,
  title = {Understanding {R1}-Zero-Like Training: A Critical Perspective},
  author = {Liu, Zichen and Chen, Changyu and Li, Wenjun and Qi, Penghui and Pang, Tianyu and Du, Chao and Lee, Wee Sun and Lin, Min},
  year = {2025},
  eprint = {2503.20783},
  archivePrefix = {arXiv},
  primaryClass = {cs.LG},
  url = {https://arxiv.org/abs/2503.20783}
}

@misc{lin2025awpo,
  title = {{AWPO}: Enhancing Tool-Use of Large Language Models through Adaptive Integration of Reasoning Rewards},
  author = {Lin, Zihan and Wang, Xiaohan and Yang, Hexiong and Chai, Jiajun and Cao, Jie and Yin, Guojun and Lin, Wei and He, Ran},
  year = {2025},
  eprint = {2512.19126},
  archivePrefix = {arXiv},
  primaryClass = {cs.CL},
  url = {https://arxiv.org/abs/2512.19126}
}

@misc{zhang2026fissiongrpo,
  title = {Robust Tool Use via {Fission-GRPO}: Learning to Recover from Execution Errors},
  author = {Zhang, Zhiwei and Zhao, Fei and Wang, Rui and Wang, Zezhong and Liang, Bin and Wang, Jiakang and Hu, Yao and Cao, Shaosheng and Wong, Kam-Fai},
  year = {2026},
  eprint = {2601.15625},
  archivePrefix = {arXiv},
  primaryClass = {cs.CL},
  url = {https://arxiv.org/abs/2601.15625}
}

@inproceedings{li2025emo,
  title = {{EMO}-{RL}: Emotion-Rule-Based Reinforcement Learning Enhanced Audio-Language Model for Generalized Speech Emotion Recognition},
  author = {Li, Pengcheng and Zhao, Botao and Kang, Zuheng and Peng, Junqing and Qu, Xiaoyang and He, Yayun and Wang, Jianzong},
  booktitle = {Findings of the Association for Computational Linguistics: EMNLP 2025},
  pages = {18744--18754},
  year = {2025},
  address = {Suzhou, China},
  publisher = {Association for Computational Linguistics},
  doi = {10.18653/v1/2025.findings-emnlp.1018},
  url = {https://aclanthology.org/2025.findings-emnlp.1018/}
}

@misc{zhang2026pointcot,
  title = {{PointCoT}: A Multi-modal Benchmark for Explicit 3D Geometric Reasoning},
  author = {Zhang, Dongxu and Sun, Yiding and Li, Pengcheng and Liu, Yumou and Lin, Hongqiang and Xu, Haoran and Mu, Xiaoxuan and Lin, Liang and Yan, Wenbiao and Yang, Ning and Fang, Chaowei and Zhao, Juanjuan and Zhu, Jihua and He, Conghui and Tan, Cheng},
  year = {2026},
  eprint = {2602.23945},
  archivePrefix = {arXiv},
  primaryClass = {cs.CV},
  url = {https://arxiv.org/abs/2602.23945}
}

@misc{sun2026tri,
  title = {Tri-Efficient Transfer Learning for Point Cloud Videos},
  author = {Sun, Yiding and Zhang, Dongxu and Zhu, Jihua and Cheng, Haozhe and Li, Zhengqiao and Li, Pengcheng and Fang, Chaowei and Dong, Yonghao and Chen, Lin},
  year = {2026},
  eprint = {2606.24175},
  archivePrefix = {arXiv},
  primaryClass = {cs.CV},
  url = {https://arxiv.org/abs/2606.24175}
}

@misc{zhang2026spark,
  title = {{SPARK}: Susceptibility-Guided Profiling and Steering of Latent Reasoning States in Large Language Models},
  author = {Zhang, Dongxu and Sun, Yiding and Guo, Zihao and Yang, Xiangyang and Tang, Kai and Chen, Lin and Tan, Cheng and Zhu, Jihua},
  year = {2026},
  eprint = {2607.10296},
  archivePrefix = {arXiv},
  primaryClass = {cs.AI},
  url = {https://arxiv.org/abs/2607.10296}
}

\appendix
\setcounter{topnumber}{4}
\setcounter{dbltopnumber}{3}
\setcounter{totalnumber}{6}
\renewcommand{\topfraction}{0.95}
\renewcommand{\dbltopfraction}{0.95}
\renewcommand{\textfraction}{0.05}
\renewcommand{\floatpagefraction}{0.85}
\renewcommand{\dblfloatpagefraction}{0.85}

\section{Training Algorithm Details}
\label{app:complete_procedure}

Algorithm~\ref{alg:vict} outlines the detailed execution flow of \textsc{VICT}. The procedure follows the same rollout and policy-update loop as the base group optimizer, but inserts verifier tracing before the advantage tensor is passed to the clipped policy-gradient objective. Verifier atoms and proof logs are never included in the policy prompt or exposed at inference time.

\begin{algorithm}[t]
\caption{Detailed Training Procedure of \textsc{VICT}}
\label{alg:vict}
\small
\begin{algorithmic}[1]
\Require Policy $\pi_\theta$, reference policy $\pi_{\mathrm{ref}}$, task distribution $\mathcal D$
\Require Group size $N$, interface builder $\mathrm{BuildInterface}$, tolerance $\eta_x$
\Require Tie margin $\epsilon_q$, core threshold $\rho_i$, budget $B$, relation weights $\{\beta_r\}$, scale $\lambda$, clip $c$
\For{iteration $k=1,\ldots,K$}
    \State Set $\theta_{\mathrm{old}}\leftarrow\theta$ and sample a task batch $x\sim\mathcal D$
    \State Collect rollouts $\{\tau_i\}_{i=1}^{N}\sim\pi_{\theta_{\mathrm{old}}}(\cdot|x)$ with the base prompt and rollout budget
    \State Evaluate $R_i=V_x(\tau_i)$ and compute base advantage $A_i^{\mathrm{base}}$
    \State Build or retrieve $\mathcal I_x=(\mathcal Z_x,F_x,D_x,M_x,\Phi_x,C_x)$
    \State Evaluate temporal atom traces $\mathbf z_i=(z_{i,j,t})_{j,t}$
    \State Run reconstruction and mutation conformance; mask failed atom families
    \State Construct proof graph $G_i=(A_i,Z_i,E_i)$ from write, reveal, commit, and violation witnesses
    \State Set $q_i$ by the sign of $A_i^{\mathrm{base}}$ using Eq.~\ref{eq:base_adv_direction}; abstain when $q_i=0$
    \State Run dependency-core search for $C_i^\star$ with budget $B$ and threshold $\rho_i$
    \If{no dependency-valid core reaches $\rho_i$}
        \State Set all verifier corrections for $\tau_i$ to zero
    \Else
        \State Compute leave-one-out marginals $\delta_{i,j}$ and group-normalize supported atoms
        \State Redistribute atom marginals to proof-bearing actions with Eq.~\ref{eq:proof_weighted_adv}
    \EndIf
    \State Form $A_{i,t}^{\mathrm{final}}=A_i^{\mathrm{base}}+\lambda\,\mathrm{clip}(A^{\mathrm{VICT}}_{i,t},-c,c)$
    \State Update $\pi_\theta$ using the base clipped policy-gradient objective and KL regularization
    \State Log each non-zero correction with action, atom, witness, evidence source, marginal, weight, and proof tag
\EndFor
\end{algorithmic}
\end{algorithm}

\paragraph{Training configuration.}
Table~\ref{tab:training_config} lists the \textsc{VICT} configuration for ALFWorld and WebShop; $\tau$-bench uses Qwen3-8B under simulated-user interaction.

\begin{table}[!t]
\centering
\small
\setlength{\tabcolsep}{4.5pt}
\renewcommand{\arraystretch}{1.05}
\resizebox{\columnwidth}{!}{
\begin{tabular}{lcc}
\toprule
\textbf{Setting} & \textbf{ALFWorld} & \textbf{WebShop} \\
\midrule
Backbones & Qwen2.5-1.5B/7B & Qwen2.5-1.5B/7B \\
Group size & 8 & 8 \\
Max episode steps & 50 & 15 \\
Max prompt length & 2048 & 4096 \\
Max response length & 512 & 512 \\
Actor learning rate & $1\times 10^{-6}$ & $1\times 10^{-6}$ \\
KL coefficient & 0.01 & 0.01 \\
Rollout temperature & 1.0 & 1.0 \\
Evaluation temperature & 0.4 & 0.4 \\
Clip ratio & 0.2 & 0.2 \\
\bottomrule
\end{tabular}
}
\caption{\textsc{VICT} training configuration for ALFWorld and WebShop.}
\label{tab:training_config}
\end{table}

\paragraph{Prompt handling and statistical reporting.}
The base agent prompt follows the ReAct-style format used by the baselines: the policy observes the task instruction, recent history, current observation, and admissible actions, then emits a reasoning block and one action. \textsc{VICT} performs evidence extraction outside the policy context by deterministic adapters over logs, state diffs, observations, and verifier metadata. The \textsc{VICT} rows report mean and standard deviation over three random seeds.

\begin{table*}[!tbp]
\centering
\small
\setlength{\tabcolsep}{4.0pt}
\renewcommand{\arraystretch}{1.05}
\begin{tabular}{lcccccc}
\toprule
\textbf{Benchmark} & \textbf{Task groups/update} & \textbf{Eval tasks} & \textbf{Updates} & \textbf{Rollouts/update} & \textbf{Train rollouts/seed} & \textbf{Checkpoint rule} \\
\midrule
ALFWorld & 16 & 134 & 300 & 128 & 38.4k & last-five average \\
WebShop & 16 & 500 & 300 & 128 & 38.4k & last-five average \\
\bottomrule
\end{tabular}
\caption{Training scale used for \textsc{VICT} on ALFWorld and WebShop. Each task group contains 8 rollouts.}
\label{tab:protocol_scale}
\end{table*}

Because online RL samples training prompts repeatedly, Table~\ref{tab:protocol_scale} reports task groups and rollouts per update rather than treating training as a single pass over a static dataset.

\section{Additional Experimental Results}
\label{app:additional_results}

\subsection{Sample-Efficiency Summary}
\label{app:sample_efficiency}

This section summarizes final scores and normalized validation AUC over the same 300 updates.

\begin{table}[!htbp]
\centering
\small
\setlength{\tabcolsep}{4.4pt}
\renewcommand{\arraystretch}{1.03}
\resizebox{\columnwidth}{!}{
\begin{tabular}{lcccc}
\toprule
\textbf{Method} & \textbf{ALF final} & \textbf{ALF AUC} & \textbf{Web final} & \textbf{Web AUC} \\
\midrule
GRPO & 77.6\textsubscript{\textpm5.2} & 58.4\textsubscript{\textpm3.8} & 66.1\textsubscript{\textpm3.7} & 52.7\textsubscript{\textpm2.9} \\
Best non-\textsc{VICT} & 91.4\textsubscript{\textpm2.3} & 71.8\textsubscript{\textpm2.1} & 76.2\textsubscript{\textpm3.4} & 61.4\textsubscript{\textpm2.6} \\
Dense atom reward & 88.6\textsubscript{\textpm1.7} & 72.6\textsubscript{\textpm1.9} & 78.2\textsubscript{\textpm1.5} & 64.9\textsubscript{\textpm1.8} \\
\textsc{VICT} & \textbf{93.7}\textsubscript{\textpm0.8} & \textbf{78.9}\textsubscript{\textpm1.4} & \textbf{83.6}\textsubscript{\textpm0.9} & \textbf{70.8}\textsubscript{\textpm1.2} \\
\bottomrule
\end{tabular}
}
\caption{Sample-efficiency summary over 300 training updates. AUC is computed from validation success curves and normalized to the same percentage scale as final performance. The best non-\textsc{VICT} row uses HCAPO for ALFWorld and SALT for WebShop.}
\label{tab:sample_efficiency}
\end{table}

\section{Verifier Interface and Domain Instantiations}
\label{app:verifier_interface_construction}

\paragraph{Atom schema.}
Each verifier atom stores a stable id, executable predicate, evaluability condition, dependency metadata, state/evidence variables, irreversibility flag, score role, and audit source. The executable predicate is the source of truth; natural-language descriptions are only metadata for inspection. This schema keeps atom decomposition stricter than a natural-language rubric because an atom must be executable, evidence-backed, or explicitly terminal-only.

\begin{table*}[!tbp]
\centering
\small
\setlength{\tabcolsep}{3.8pt}
\renewcommand{\arraystretch}{1.04}
\begin{tabular}{p{0.16\textwidth}p{0.36\textwidth}p{0.40\textwidth}}
\toprule
\textbf{Field} & \textbf{Meaning} & \textbf{Examples} \\
\midrule
\texttt{atom\_id} & Stable identifier for the checked fact. & \texttt{webshop.price\_under\_budget}, \texttt{tau.confirmation\_obtained}. \\
\texttt{predicate} & Executable function mapping trajectory state/evidence to $\{\mathrm{sat},\mathrm{unsat},\mathrm{unk},\mathrm{viol}\}$. & Product price comparison; database-field equality; ALFWorld object-state predicate. \\
\texttt{evaluable\_when} & Earliest point at which the atom can be safely evaluated. & Online state, evidence revealed, commit action, or terminal only. \\
\texttt{dependencies} & Parent atoms or clauses required for valid counterfactual edits. & ``field update allowed'' depends on ``confirmation obtained''. \\
\texttt{state\_vars} & Variables used by $M_x$ and $\Phi_x$ to attach witnesses. & Product id, selected option, object location, database record field. \\
\texttt{score\_role} / \texttt{audit\_source} & Effect on $F_x$ and evidence saved for audit. & Hard conjunction, penalty, prerequisite; state diff, observation span, API log. \\
\bottomrule
\end{tabular}
\caption{Verifier atom schema used by \textsc{VICT}.}
\label{tab:atom_schema}
\end{table*}

\paragraph{Three-level instrumentation.}
Level 0 creates atoms from exposed final-state or goal-state differences, such as field equality, record existence, and forbidden non-modification. Level 1 instruments explicit verifier branches, assertions, policy checks, or score terms when their truth values can be recomputed from the trajectory, terminal state, or logged evidence. Level 2 uses small deterministic adapters to expose verifier-relevant metadata or semi-structured observations; such adapters may only expose facts already used by the terminal verifier or required to reproduce it.

\begin{figure}[t]
\centering
\includegraphics[width=\columnwidth]{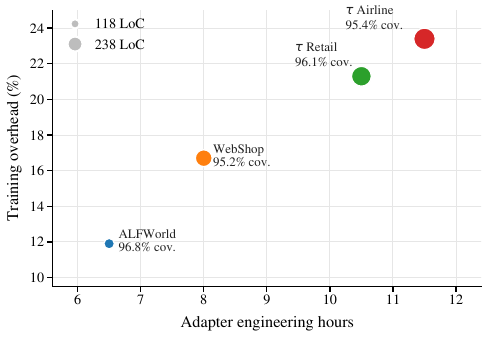}
\caption{Adapter engineering cost and training-time overhead across domains. Marker size encodes adapter lines of code, and labels report atom coverage.}
\label{fig:cost_overhead}
\end{figure}

\paragraph{Domain instantiations.}
Table~\ref{tab:domain_instantiations} summarizes example atom families, witness sources, and mutation tests. ALFWorld relies heavily on simulator state diffs, WebShop relies on product evidence and purchase commits, and $\tau$-bench uses API logs, database diffs, user turns, and policy checks.

\begin{table*}[!tbp]
\centering
\small
\setlength{\tabcolsep}{3.8pt}
\renewcommand{\arraystretch}{1.06}
\begin{tabular}{p{0.12\textwidth}p{0.29\textwidth}p{0.29\textwidth}p{0.22\textwidth}}
\toprule
\textbf{Domain} & \textbf{Atom families} & \textbf{Witness sources} & \textbf{Mutation tests} \\
\midrule
ALFWorld &
Target object identity; object held; transformation; receptacle found; final placement; second-object requirement; wrong-object or premature-placement violation. &
Simulator state diffs; inventory changes; temperature or cleanliness transitions; receptacle transitions; final placement actions. &
Move final object to wrong receptacle; remove transformation flag; swap target object; drop second object. \\
\midrule
WebShop &
Product identity; category match; hard attribute match; option selected; price under budget; purchase committed; wrong-purchase violation. &
Search results; product page metadata; option-click logs; selected product state; final buy action. &
Change price; remove required attribute; change selected option; replace purchased product id. \\
\midrule
$\tau$-bench &
Target record identified; required information collected; confirmation obtained; policy precondition; target field updated; forbidden field unchanged; final response consistent. &
API call logs; database diffs; user turns; policy-rule checks; final assistant response. &
Delete confirmation; alter target field; add extra database write; change policy precondition. \\
\bottomrule
\end{tabular}
\caption{Example \textsc{VICT} instantiations.}
\label{tab:domain_instantiations}
\end{table*}

\section{Attribution, Core Search, and Normalization}
\label{app:attribution_details}

\paragraph{Conformance tests.}
An interface is accepted only when it reconstructs the original terminal verifier within a fixed tolerance. For held-out rollouts and verifier-relevant mutations $\mathcal S_x=\mathcal T_x\cup\mathfrak M_x(\mathcal T_x)$, we compute
\begin{equation}
e_x(\tau)=
\left|F_x(\bar{\mathbf z}(\tau))-V_x(\tau)\right|.
\end{equation}
Reward reconstruction and mutation conformance are the fractions of examples whose error is at most $\eta_x$. For exact Boolean verifiers we set $\eta_x=0$; for graded scores, $\eta_x$ is fixed before RL training. If one atom family fails conformance while the rest reconstructs correctly, only that family is masked; if the aggregator fails, all \textsc{VICT} corrections for the task are zeroed.

\paragraph{Witness relations and proof gates.}
\textsc{VICT} creates action-to-atom edges only through fixed witness predicates: direct writes, evidence reveals, commits, and violations. A direct-write relation is created when an action changes a verifier-relevant variable mapped to an atom:
\[
\begin{aligned}
W_{\mathrm{write}}(i,t,j)=
\mathbf 1\{&\exists v\in M_x(z_j):\\
&\Phi_x(h_{i,t+1})[v]\neq \Phi_x(h_{i,t})[v]\}.
\end{aligned}
\]
Evidence-reveal edges credit actions that make previously unavailable verifier evidence visible; commit edges attach finalizing actions such as \texttt{buy}, answer submission, database update, or object placement; violation edges attach irreversible forbidden changes. Terminal-only atoms are attached only when a reliable last writer or final commit action exists.

Every non-zero correction is emitted only if the proof gate passes:
\begin{equation}
\begin{aligned}
\mathrm{Gate}_{i,t,j,r}=1
&\Rightarrow
\epsilon_x^{\mathrm{conf}}\leq \eta_x,\\
&\quad j\in C_i^\star,\quad
(t,j,r)\in E_i,\\
&\quad \delta_{i,j}\neq 0,\quad
\mathcal P_{i,t,j}\models D_x .
\end{aligned}
\end{equation}
Proof records store the action id, atom id, witness type, evidence source, core id, marginal, witness weight, and assigned correction.

\paragraph{Eligibility-soundness proof sketch.}
\label{app:eligibility_proof}
Eq.~\ref{eq:proof_weighted_adv} can be non-zero only when $\hat{\delta}_{i,j}\neq0$ and $\omega_{i,t,j}>0$ for a core atom. By construction, this requires verifier conformance, membership in $C_i^\star$, an observed witness edge, a non-zero marginal, and a proof record satisfying $D_x$. These are exactly the conditions summarized by $\mathrm{Gate}_{i,t,j,r}=1$ in Eq.~\ref{eq:proof_carrying_invariant}.

\paragraph{Budgeted greedy search.}
\label{app:budgeted_greedy_search}
The greedy core search prioritizes validity and auditability over global optimality. Starting from $C^{(0)}=\emptyset$, at iteration $\ell$ it selects
\begin{equation}
\begin{aligned}
j_\ell
&=
\operatorname*{arg\,max}_{j\notin C^{(\ell)}}
\Delta_x(\mathrm{cl}_{D_x}(C^{(\ell)}\cup\{j\});
\bar{\mathbf z}_i,q_i),\\
C^{(\ell+1)}
&=
\mathrm{cl}_{D_x}(C^{(\ell)}\cup\{j_\ell\}).
\end{aligned}
\end{equation}
The search stops when $\Delta_x(C^{(\ell)};\bar{\mathbf z}_i,q_i)\geq\rho_i$ or when budget $B$ is exhausted. A backward deletion pass removes any atom whose deletion preserves both closure and the displacement threshold. We do not claim a global approximation ratio: a suboptimal search can reduce credit recall or return a larger locally minimal core, but the proof-edge eligibility invariant is preserved. Ties are broken by a fixed atom index, so the search is deterministic for a fixed trace.

\paragraph{Counterfactual validity and cost.}
The dependency closure $\mathrm{cl}_{D_x}$ is implemented as the least fixed point of the verifier dependency rules; ambiguous closures are marked unsupported. Counterfactual scores use $F_x$, and bounded conformance error implies
\begin{equation}
\left|\Delta_x^{F}(C)-\Delta_x^{V}(C)\right|\leq 2\eta_x .
\end{equation}
Cores inside $[\rho_i-2\eta_x,\rho_i+2\eta_x]$ are abstained unless exact conformance is available. The per-rollout overhead is
\begin{equation}
O(T_i m |\mathcal R_x| c_W + Bm c_F),
\end{equation}
where the first term constructs proof edges and the second evaluates candidate dependency cores.

\paragraph{Robust normalization and abstention.}
\label{app:robust_normalization}
The robust normalization is computed over the current rollout group for the same task instance:
\begin{equation}
\operatorname{Norm}_{x,j}(\delta_{i,j})
=
\frac{\delta_{i,j}-\mathrm{med}_{x,j}}{s_{x,j}+\epsilon}.
\label{eq:atom_norm}
\end{equation}
If fewer than two supported rollouts contain atom $j$, or if the robust scale is zero, the normalized marginal is set to zero. Abstention is implemented by zeroing the eligible core signal for conformance failures, near ties, failed core search, zero robust scale, or missing proof support; in all cases, $A_{i,t}^{\mathrm{final}}=A_i^{\mathrm{base}}$ when all verifier corrections are zero.

\paragraph{Policy-invariance and reward-shaping scope.}
\label{app:policy_invariance_scope}
\textsc{VICT} does not claim the policy-invariance guarantee of potential-based reward shaping. The terminal verifier $V_x$ remains the outcome target, but Eq.~\ref{eq:final_adv} changes the stochastic gradient update through a clipped verifier-derived correction. The correction can introduce useful bias through $\lambda$, clipping, greedy core search, dependency design, and proof coverage. The safeguards are operational: corrections are clipped, missing or ambiguous evidence abstains to $A_i^{\mathrm{base}}$, and every non-zero correction is logged through the verifier interface. The empirical question is whether this verifier-grounded bias improves learning under matched rollout budgets.

\begin{table}[!t]
\centering
\small
\setlength{\tabcolsep}{5.0pt}
\renewcommand{\arraystretch}{1.05}
\resizebox{\columnwidth}{!}{
\begin{tabular}{lc}
\toprule
\textbf{Hyperparameter} & \textbf{Default} \\
\midrule
Core budget $B$ & 8 atom additions \\
Core threshold $\rho_i$ & $0.5\,s_R|A_i^{\mathrm{base}}|$ \\
Tie margin $\epsilon_q$ for $q_i$ & $10^{-6}$ base-advantage units \\
Conformance tolerance $\eta_x$ & 0 for exact verifiers \\
Direct-write / commit / violation weight & 1.0 \\
Evidence-path total weight & 1.0 split over support path \\
Correction scale $\gamma_\lambda$ & 0.5 \\
Numerical $\epsilon$ & $10^{-6}$ \\
\bottomrule
\end{tabular}
}
\caption{Default values for the verifier correction.}
\label{tab:vict_hyperparameters}
\end{table}

Here $s_R$ is the current rollout group's reward scale used by the base optimizer. For standardized GRPO-style advantages, $s_R|A_i^{\mathrm{base}}|$ recovers the unnormalized preference magnitude in verifier-score units; for RLOO-style advantages, the same principle uses the reward scale implicit in the base advantage.

The correction scale is calibrated to the base advantage scale:
\begin{equation}
\begin{aligned}
\lambda
\ &=\
\gamma_\lambda
\frac{\mathrm{median}_{i,t}|A_i^{\mathrm{base}}|}
{\mathrm{median}_{i,t}|A^{\mathrm{VICT}}_{i,t}|+\epsilon},\\
c=q_{0.95}(|A_i^{\mathrm{base}}|).
\end{aligned}
\end{equation}

\section{Diagnostics and Ablation Interpretation}
\label{app:diagnostics}

\paragraph{Faithfulness metrics.}
We report reward reconstruction, mutation conformance, eligibility-invariant pass rate, proof coverage, abstention rate, and credit sparsity. The eligibility-invariant pass rate is the fraction of non-zero corrections satisfying Eq.~\ref{eq:proof_carrying_invariant}; it is primarily an implementation invariant, so a non-zero violation rate indicates an error rather than a meaningful empirical tradeoff. Let $p_{i,j}=\mathbf 1\{\delta_{i,j}\neq0,\exists t,r:(t,j,r)\in E_i\}$. Proof coverage is
\[
\mathrm{ProofCov}=
\frac{\sum_{i,j}p_{i,j}}
{\sum_{i,j}\mathbf 1\{\delta_{i,j}\neq 0\}+\epsilon}.
\]
Credit sparsity is the fraction of trajectory actions with $|A^{\mathrm{VICT}}_{i,t}|>0$. We separately log abstention from conformance failure, near ties, core-search failure, uncertainty-band failure, zero robust scale, and missing proof support.

\paragraph{Core and hyperparameter diagnostics.}
Table~\ref{tab:credit_behavior} reports the mean and standard deviation of returned core size together with the budget-hit rate. The default $B=8$ is intended for the small per-instance atom sets in Table~\ref{tab:verifier_diagnostics}; larger atom sets should report core-size distributions and a sensitivity sweep over $B$. The correction scale $\gamma_\lambda$ is calibrated against the base advantage scale, but its sweep should be reported when verifier score ranges change substantially.

\paragraph{Ablation interpretation.}
The dense atom reward ablation tests whether gains come merely from exposing intermediate verifier facts. It does not prove that \textsc{VICT} is not return redistribution; instead, it tests whether the dependency core, proof edge, and abstention constraints add value beyond direct dense atom rewards. The no-core ablation tests whether all satisfied facts can be treated as relevant. The no-proof-edge ablation tests whether temporal proximity is enough to assign credit. The full method requires all three conditions simultaneously: an atom must affect a dependency-valid verifier core, have an observable proof edge, and survive group-normalized masking.

\paragraph{Targeted negative controls.}
\label{app:targeted_controls}
Table~\ref{tab:ablation} includes three targeted controls that probe simpler explanations for the gains. Commit-only verifier credit attaches verifier credit only to finalizing actions such as \texttt{buy}, object placement, or API update. Temporal-nearest atom credit assigns each atom marginal to the nearest preceding action with any lexical or state overlap. Randomized proof-edge placebo preserves sparsity by permuting proof edges within the same rollout group while keeping the number of credited actions fixed.

\begin{table*}[!tbp]
\centering
\small
\setlength{\tabcolsep}{3.2pt}
\renewcommand{\arraystretch}{1.05}
\begin{tabular}{lccccccc}
\toprule
\textbf{Domain} & \textbf{Trigger} & \textbf{Sparsity} & \textbf{NZ acts/traj} & \textbf{Core size} & \textbf{Budget hit} & \textbf{Abstain} & \textbf{Top abstention reason} \\
\midrule
ALFWorld & 61.8\textsubscript{\textpm2.1}\% & 18.6\textsubscript{\textpm1.0}\% & 4.7\textsubscript{\textpm0.4} & 2.9\textsubscript{\textpm0.2} & 1.4\textsubscript{\textpm0.3}\% & 8.1\textsubscript{\textpm0.6}\% & terminal-only fact \\
WebShop & 57.2\textsubscript{\textpm1.8}\% & 15.3\textsubscript{\textpm0.9}\% & 2.2\textsubscript{\textpm0.2} & 3.4\textsubscript{\textpm0.3} & 2.1\textsubscript{\textpm0.4}\% & 11.6\textsubscript{\textpm0.8}\% & ambiguous product evidence \\
$\tau$ Retail & 52.9\textsubscript{\textpm2.4}\% & 12.8\textsubscript{\textpm1.1}\% & 3.1\textsubscript{\textpm0.3} & 4.6\textsubscript{\textpm0.4} & 4.8\textsubscript{\textpm0.6}\% & 14.3\textsubscript{\textpm1.0}\% & missing confirmation witness \\
$\tau$ Airline & 49.5\textsubscript{\textpm2.6}\% & 11.7\textsubscript{\textpm1.0}\% & 3.4\textsubscript{\textpm0.4} & 5.1\textsubscript{\textpm0.5} & 5.6\textsubscript{\textpm0.7}\% & 15.8\textsubscript{\textpm1.1}\% & unresolved policy precondition \\
\bottomrule
\end{tabular}
\caption{Credit behavior diagnostics computed on final training rollouts before policy update. ALFWorld/WebShop values are averaged over three training seeds; $\tau$-bench values are averaged over final service-validation rollout batches because that setting is treated as supplemental validation. Trigger is the fraction of rollouts with at least one non-zero verifier correction; sparsity is the fraction of actions with non-zero verifier correction.}
\label{tab:credit_behavior}
\end{table*}

\section{Illustrative Proof Traces and Benchmark Case Studies}
\label{app:illustrative_traces}

This section illustrates the verifier interface with benchmark cases. The ALFWorld and WebShop examples follow their task and observation formats; the $\tau$-bench examples use released Retail and Airline task records. The sign columns are schematic: they indicate whether the illustrated evidence supports, opposes, or does not support an action-level correction before group-dependent scaling, rather than constituting additional quantitative results.

\subsection{ALFWorld: Heat an Egg and Place It on the Countertop}
\label{app:alfworld_case_study}

The ALFWorld case is a \textit{Heat} task: \textit{heat some egg and put it in countertop}. The terminal verifier checks object identity, transformation, and final placement. Table~\ref{tab:alfworld_heat_egg_trace} shows how \textsc{VICT} separates useful search, state-changing actions, and final commitment. Opening the fridge is not itself rewarded as progress when no verifier atom enters the dependency core; taking the egg, heating it, and moving the heated egg to the countertop are credited because they write verifier-relevant state.

\begin{table*}[!tbp]
\centering
\scriptsize
\setlength{\tabcolsep}{3.1pt}
\renewcommand{\arraystretch}{1.08}
\begin{tabular}{cp{0.24\textwidth}p{0.21\textwidth}p{0.23\textwidth}p{0.13\textwidth}c}
\toprule
\textbf{Step} & \textbf{Observation / action} & \textbf{Evidence used by adapter} & \textbf{Verifier atom status after action} & \textbf{Proof edge} & \textbf{Sign} \\
\midrule
0 & Initial kitchen contains \texttt{countertop 1}, \texttt{countertop 2}, \texttt{fridge 1}, and \texttt{microwave 1}. & Receptacle and heating-device candidates are visible. & \texttt{receptacle\_candidate=seen}; \texttt{heat\_tool=seen}; target object unknown. & evidence reveal & $0$ \\
1--2 & \texttt{go to fridge 1}; \texttt{open fridge 1}. & Fridge contains bowl, pan, plate, and potato, but no egg. & No target-object atom is satisfied; the negative observation is logged but not a success core atom. & reveal, no core & $0$ \\
3--4 & \texttt{go to countertop 2}; observation lists \texttt{egg 3}, \texttt{egg 2}, and \texttt{egg 1}. & The target object is now grounded to visible egg instances. & \texttt{target\_object\_visible=sat}. & evidence reveal & $+$ \\
5 & \texttt{take egg 1 from countertop 2}. & Inventory changes from no target object to holding \texttt{egg 1}. & \texttt{target\_object\_held=sat}. & direct write & $+$ \\
6--7 & \texttt{go to microwave 1}; \texttt{open microwave 1}. & The heating device is reachable and usable. & Dependency for \texttt{heated\_egg} is prepared, but no transformation has happened yet. & support path & $0/+$ \\
8 & \texttt{heat egg 1 with microwave 1}. & Simulator state marks \texttt{egg 1} as heated. & \texttt{transformation\_heated=sat}. & direct write & $+$ \\
9 & \texttt{go to countertop 1}. & Target receptacle is reached. & Placement is still unsatisfied, but the next commit scope is valid. & support path & $0/+$ \\
10 & \texttt{move egg 1 to countertop 1}. & Final location of heated \texttt{egg 1} becomes \texttt{countertop 1}. & \texttt{final\_placement=sat}; all parents satisfied. & direct write + commit & $+$ \\
\bottomrule
\end{tabular}
\caption{ALFWorld proof trace for an illustrative \textit{Heat} task. \textsc{VICT} credits the actions that write target identity, heating, and placement atoms, while search actions only receive credit when they lie on a proof-supported evidence path.}
\label{tab:alfworld_heat_egg_trace}
\end{table*}

The same interface also explains common failures more precisely than outcome-only credit. Table~\ref{tab:alfworld_heat_egg_failures} shows four near-miss continuations from the same task. A failed trajectory that finds and holds the egg is not treated as uniformly bad: the object-acquisition edge can keep positive correction, while the premature or wrong final placement receives negative commit credit.

\begin{table*}[!tbp]
\centering
\scriptsize
\setlength{\tabcolsep}{3.2pt}
\renewcommand{\arraystretch}{1.08}
\begin{tabular}{p{0.18\textwidth}p{0.28\textwidth}p{0.25\textwidth}p{0.19\textwidth}}
\toprule
\textbf{Near miss} & \textbf{Verifier diagnosis} & \textbf{Action receiving correction} & \textbf{Why this is different from outcome-only RL} \\
\midrule
Takes \texttt{potato 1} from the fridge after failing to find an egg. & \texttt{wrong\_object\_held=viol}; target egg atoms remain unsatisfied. & The wrong \texttt{take} action receives negative violation credit if it enters the failure core. & The model is not told that all early navigation was wrong; the blame is attached to the first wrong-object write. \\
Moves \texttt{egg 1} to the countertop before heating. & \texttt{target\_object\_held=sat}, but \texttt{transformation\_heated=unsat} at final placement. & \texttt{move egg 1 to countertop 1} receives negative premature-commit credit; \texttt{take egg 1} may still receive positive object credit. & A partially useful failed trajectory is decomposed into a correct object choice and an incorrect final commitment. \\
Heats the egg, then places it in the sinkbasin. & \texttt{transformation\_heated=sat}; \texttt{final\_receptacle=viol}. & The wrong \texttt{move} or \texttt{put} action receives negative placement credit. & The heating action remains supported even though the terminal reward is zero. \\
Loops through \texttt{look} and repeated container openings after the egg is held. & No additional atom changes; no reveal edge adds new verifier evidence. & Repeated no-op actions receive no verifier correction. & \textsc{VICT} does not create dense progress rewards for merely longer successful-looking histories. \\
\bottomrule
\end{tabular}
\caption{ALFWorld failure contrasts. The dependency core allows useful predecessor actions and harmful commit actions to receive different signs inside the same failed rollout.}
\label{tab:alfworld_heat_egg_failures}
\end{table*}

\subsection{WebShop: Premature Purchase with Correct Options but Wrong Product}
\label{app:webshop_case_study}

The WebShop case uses the instruction: \textit{Find me loose fit, slim fit men's tuxedo shirts with long sleeve, short sleeve, polyester cotton, elastic waist, regular fit for gym workout with color: b-blue, and size: xx-large, and price lower than 40.00 dollars}. The observed rollout searches with many requested fields, clicks product \texttt{B09Q67H373}, selects \texttt{b-blue} and \texttt{xx-large}, and then buys. The product page exposes color and size options and a price below the budget, but the title describes a T-shirt rather than the requested tuxedo shirt with the required material and fit constraints. Table~\ref{tab:webshop_tuxedo_trace} shows why this is an informative failure for \textsc{VICT}.

\begin{table*}[!tbp]
\centering
\scriptsize
\setlength{\tabcolsep}{3.0pt}
\renewcommand{\arraystretch}{1.08}
\begin{tabular}{cp{0.20\textwidth}p{0.25\textwidth}p{0.24\textwidth}p{0.13\textwidth}c}
\toprule
\textbf{Step} & \textbf{Action} & \textbf{Observation / evidence} & \textbf{Verifier atom status} & \textbf{Proof edge} & \textbf{Sign} \\
\midrule
1 & Search with color \texttt{b-blue}, size \texttt{xx-large}, budget \texttt{0-40}, product type, fit, and sleeve slots. & Search query contains color, size, budget, product type, fit, and sleeve constraints. & Query covers several hard constraints, but no product atom is yet evaluable. & reveal path & $+$ \\
2 & \texttt{click[b09q67h373]} & Result title: ``JSPOYOU 2022 Newly T-Shirt... Short Sleeve Tees...''; price \texttt{\$1.99}--\texttt{\$8.99}. & \texttt{price\_under\_budget=sat}; \texttt{product\_type\_tuxedo=unsat/unk}; material and elastic-waist evidence absent. & evidence reveal & $+/-$ \\
3 & \texttt{click[b-blue]} & Product page exposes color options including \texttt{b-blue}. & \texttt{color\_option\_selected=sat}. & direct write & $+$ \\
4 & \texttt{click[xx-large]} & Product page exposes size options including \texttt{xx-large}. & \texttt{size\_option\_selected=sat}. & direct write & $+$ \\
5 & \texttt{click[buy now]} & Final purchase commits the current product, whose page lacks the tuxedo-shirt, polyester-cotton, elastic-waist, and requested-fit evidence. & \texttt{purchase\_committed=sat}; hard product atoms missing. & commit violation & $-$ \\
\bottomrule
\end{tabular}
\caption{WebShop proof trace for an illustrative failed purchase. Option-selection actions can receive positive correction, but the final purchase receives negative commit credit because dependency-core atoms for product type and hard attributes remain unsatisfied.}
\label{tab:webshop_tuxedo_trace}
\end{table*}

Table~\ref{tab:webshop_tuxedo_contrasts} shows the behavior that this proof trace encourages. The model should not learn only ``buy after selecting color and size''; it should learn to delay purchase until product-page evidence supports all hard atoms. This is where \textsc{VICT} differs from dense atom reward: a page that satisfies cheap price and selectable options is still not a positive terminal core if the purchased product violates the hard instruction.

\begin{table*}[!tbp]
\centering
\scriptsize
\setlength{\tabcolsep}{3.2pt}
\renewcommand{\arraystretch}{1.08}
\begin{tabular}{p{0.21\textwidth}p{0.26\textwidth}p{0.24\textwidth}p{0.19\textwidth}}
\toprule
\textbf{Continuation} & \textbf{Satisfied atoms} & \textbf{Missing or violated atoms} & \textbf{\textsc{VICT} effect} \\
\midrule
Select \texttt{b-blue} and \texttt{xx-large}, then buy \texttt{B09Q67H373}. & Price, color option, and size option are supported by page evidence. & Product category, material, elastic waist, and requested fit are not supported. & Positive option writes are preserved, but the \texttt{buy now} action is penalized as a wrong purchase. \\
Click a cheap, semantically clothing-related result without checking product details. & Search result may reveal a low price and some lexical overlap. & Hard attributes are unknown, so the core cannot certify purchase correctness. & Evidence is insufficient; \textsc{VICT} either abstains or credits only the evidence-reveal edge, not a commit. \\
Return to search after detecting missing product-type evidence. & The failed candidate contributes negative evidence for the current product. & No terminal purchase is committed. & The click can be treated as useful evidence gathering rather than as a failed purchase. \\
Buy after all hard product atoms and required options are observed. & Product type, hard attributes, options, budget, and commit are all satisfied. & No missing hard atom remains in the dependency core. & Evidence-revealing clicks, option writes, and the final commit all receive positive correction. \\
\bottomrule
\end{tabular}
\caption{WebShop continuation contrasts. The decisive distinction is whether the final \texttt{buy} action commits a product whose hard verifier atoms are all supported.}
\label{tab:webshop_tuxedo_contrasts}
\end{table*}

\subsection{$\tau$-Bench Retail: Multi-Item Exchange with Fallback Preferences}
\label{app:tau_retail_case_study}

The Retail case is a released $\tau$-bench task involving a delivered order and an exchange of a mechanical keyboard and a smart thermostat. The order record contains keyboard item \texttt{1151293680} with \texttt{linear/RGB/full size} options and thermostat item \texttt{4983901480} with \texttt{Apple HomeKit/black}. The target exchange is keyboard item \texttt{7706410293} (\texttt{clicky/no backlight/full size}) and thermostat item \texttt{7747408585} (\texttt{Google Assistant/black}), paid with \texttt{credit\_card\_9513926}. The keyboard choice is subtle: the exact \texttt{clicky/RGB/full size} variant exists as item \texttt{9025753381}, but it is unavailable, so the user's fallback preference selects the available no-backlight full-size variant.

\begin{table*}[!tbp]
\centering
\scriptsize
\setlength{\tabcolsep}{2.9pt}
\renewcommand{\arraystretch}{1.08}
\begin{tabular}{cp{0.22\textwidth}p{0.26\textwidth}p{0.23\textwidth}p{0.12\textwidth}c}
\toprule
\textbf{Step} & \textbf{Action / turn} & \textbf{Evidence source} & \textbf{Verifier atom status} & \textbf{Proof edge} & \textbf{Sign} \\
\midrule
1 & Name+zip authentication via \texttt{find\_user\_id\_by\_name\_zip}. & Tool returns the matching benchmark user record. & \texttt{identity\_authenticated=sat}. & evidence reveal & $+$ \\
2 & Retrieve order \texttt{\#W2378156} with \texttt{get\_order\_details}. & Order status is \texttt{delivered}; target item ids and original payment method are visible. & \texttt{delivered\_order=sat}; target items identified. & evidence reveal & $+$ \\
3 & Check keyboard product \texttt{1656367028}. & Keyboard variants show \texttt{9025753381} is exact but unavailable; \texttt{7706410293} is clicky, no-backlight, full-size, and available. & Keyboard replacement valid; fallback condition satisfied. & evidence reveal & $+$ \\
4 & Check thermostat product \texttt{4896585277}. & Thermostat variant \texttt{7747408585} is Google Assistant compatible, black, and available. & Thermostat replacement valid. & evidence reveal & $+$ \\
5 & Assistant lists both exchanges and obtains explicit user ``yes''. & Retail policy requires confirmation before any database-changing exchange. & \texttt{confirmation\_obtained=sat}. & user-turn reveal & $+$ \\
6 & Call the exchange API with order \texttt{\#W2378156}, old ids \texttt{1151293680/4983901480}, new ids \texttt{7706410293/7747408585}, and \texttt{credit\_card\_9513926}. & Database changes status to \texttt{exchange requested} and records exact old/new item lists. & \texttt{exchange\_status=sat}; \texttt{no\_extra\_write=sat}; all item atoms satisfied. & direct write + commit & $+$ \\
\bottomrule
\end{tabular}
\caption{$\tau$-bench Retail proof trace for an illustrative multi-item exchange task. The product-detail calls reveal the unavailable exact keyboard variant, the valid fallback, and the valid thermostat replacement.}
\label{tab:tau_retail_exchange_trace}
\end{table*}

\begin{table*}[!tbp]
\centering
\scriptsize
\setlength{\tabcolsep}{3.2pt}
\renewcommand{\arraystretch}{1.08}
\begin{tabular}{p{0.24\textwidth}p{0.25\textwidth}p{0.24\textwidth}p{0.17\textwidth}}
\toprule
\textbf{Failure continuation} & \textbf{Failed atom} & \textbf{Credited or penalized action} & \textbf{Training signal} \\
\midrule
Calls the exchange tool before obtaining explicit user confirmation. & \texttt{confirmation\_obtained=unsat}; \texttt{policy\_precondition=viol}. & The mutating API call receives negative violation credit. & Learn to ask for confirmation before database writes. \\
Chooses exact keyboard item \texttt{9025753381}. & \texttt{replacement\_available=viol}; tool would reject unavailable item. & The product-detail evidence is useful, but the exchange call is penalized. & Learn that an unavailable exact variant triggers the fallback branch. \\
Chooses thermostat item \texttt{4953074738}. & \texttt{thermostat\_compatibility=viol} because it is Amazon Alexa, not Google Assistant. & The wrong new-item id in the exchange call receives negative commit credit. & Learn to bind replacement ids to checked option atoms. \\
Exchanges only the thermostat and omits the keyboard in this task variant. & \texttt{keyboard\_exchange\_included=unsat}. & The final exchange call receives omission credit for the missing item id. & Learn to batch all requested delivered-order exchanges into one mutating call. \\
\bottomrule
\end{tabular}
\caption{$\tau$-bench Retail failure contrasts. The verifier core localizes whether the error is a policy violation, an unavailable item, a wrong replacement option, or an incomplete database update.}
\label{tab:tau_retail_exchange_failures}
\end{table*}

This task illustrates why policy and database atoms must be part of the verifier core. Table~\ref{tab:tau_retail_exchange_failures} lists failure modes that a scalar terminal reward would collapse into the same zero outcome. \textsc{VICT} distinguishes missing confirmation, unavailable replacements, wrong preference resolution, and incomplete exchange lists.

\subsection{$\tau$-Bench Airline: Same-Day Return Change and Baggage Update}
\label{app:tau_airline_case_study}

\begin{table*}[!tbp]
\centering
\scriptsize
\setlength{\tabcolsep}{2.8pt}
\renewcommand{\arraystretch}{1.08}
\begin{tabular}{cp{0.21\textwidth}p{0.27\textwidth}p{0.23\textwidth}p{0.12\textwidth}c}
\toprule
\textbf{Step} & \textbf{Action / evidence} & \textbf{Database or search fact} & \textbf{Verifier atom status} & \textbf{Proof edge} & \textbf{Sign} \\
\midrule
1 & Retrieve user and reservation details. & \texttt{OBUT9V}: IAH--DEN round trip, economy; outbound is \texttt{HAT078}/\texttt{HAT118} on May 27 and return is \texttt{HAT084}/\texttt{HAT266} on May 28. & Target reservation identified; economy booking is modifiable. & evidence reveal & $+$ \\
2 & Search return flights from DEN to IAH on May 27. & The task record specifies \texttt{HAT290} DEN--LAS 14:00--16:00 and \texttt{HAT175} LAS--IAH 17:00--20:00. & \texttt{same\_day\_return=sat}; \texttt{return\_route\_selected=sat}. & evidence reveal & $+$ \\
3 & Compare payment methods. & \texttt{gift\_card\_7480005} has \$6; \texttt{gift\_card\_6276644} has \$113 and is the smallest sufficient gift card. & \texttt{smallest\_usable\_gift\_card=sat}. & evidence reveal & $+$ \\
4 & Obtain explicit user confirmation for flight and baggage changes. & Airline policy requires confirmation before updating flights or baggage. & \texttt{confirmation\_obtained=sat}. & user-turn reveal & $+$ \\
5 & Flight-update API call for \texttt{OBUT9V}: economy itinerary \texttt{HAT078/HAT118/HAT290/HAT175}, paid with \texttt{gift\_card\_6276644}. & The whole new reservation is written, retaining outbound segments and replacing the return with same-day flights. & \texttt{flight\_update=sat}; outbound preserved. & direct write + commit & $+$ \\
6 & Baggage-update API call for \texttt{OBUT9V}: \texttt{total=2}, \texttt{nonfree=0}, payment \texttt{gift\_card\_6276644}. & Total checked bags becomes 2; silver economy allowance keeps nonfree bags at 0. & \texttt{baggage\_update=sat}; free allowance respected. & direct write + commit & $+$ \\
\bottomrule
\end{tabular}
\caption{$\tau$-bench Airline proof trace for an illustrative reservation-modification task. The proof core combines itinerary search, payment sufficiency, user confirmation, and two database writes.}
\label{tab:tau_airline_change_trace}
\end{table*}

The Airline case is a released reservation-modification task. The user wants to modify reservation \texttt{OBUT9V}: keep the Houston-to-Denver outbound trip on May 27, change the return to the fastest same-day return, stay in economy, add one checked bag, and pay with the smallest usable gift card. The reservation record shows the outbound segments \texttt{HAT078} and \texttt{HAT118}, an original next-day return \texttt{HAT084}/\texttt{HAT266}, one checked bag, and no insurance. The user is a silver member, so two economy checked bags are free. The profile contains gift cards with balances 157, 113, and 6 dollars; the 6-dollar card is too small for the flight price difference, so the smallest usable card is \texttt{gift\_card\_6276644} with 113 dollars.

The Airline task also shows the value of dependency-aware attribution. Calling the flight-update API with only the return legs omits unchanged outbound segments even if the desired return is correct; using the 6-dollar gift card violates the payment sufficiency atom; setting \texttt{nonfree\_baggages=1} contradicts the silver-member free-baggage rule; and any mutating call before confirmation violates the policy precondition. \textsc{VICT} assigns the negative signal to the specific API call that writes the inconsistent field, while still crediting earlier evidence-gathering calls that identified the correct reservation, route, or payment candidate.

\end{document}